\documentclass[runningheads]{llncs}

\usepackage{eccv}

\usepackage{eccvabbrv}

\usepackage{graphicx}
\usepackage{booktabs}

\usepackage{algorithm}
\usepackage{algorithmic}
\usepackage{amsmath}
\usepackage{amssymb}
\usepackage{mathtools}

\usepackage{multirow}
\usepackage{pifont}
\newcommand{\cmark}{\ding{51}}%
\usepackage[accsupp]{axessibility}  

\usepackage{hyperref}
\hypersetup{
 colorlinks=true,
 linkcolor=red,
 citecolor=cyan,
 filecolor=magenta, 
 urlcolor=cyan,
 } 

\usepackage{orcidlink}

\def\eg{\emph{e.g.,~}}
\def\ournet{MonoTTA}

\begin{document}

\title{Fully Test-Time Adaptation for Monocular 3D Object Detection} 

\titlerunning{MonoTTA}

\author{
Hongbin Lin\inst{1}$^*$,
Yifan Zhang\inst{2}$^*$,
Shuaicheng Niu\inst{3}\thanks{Authors contributed equally.},
Shuguang Cui\inst{1},
Zhen Li\inst{1}\thanks{Corresponding authors.}
}
\authorrunning{Hongbin Lin, Yifan Zhang and et al.}
%
\institute{$^1$Chinese University of Hong Kong, Shenzhen $^2$ National University of Singapore \\
$^3$ Nanyang Technological University \\
\email{hongbinlin@link.cuhk.edu.cn}\texttt{,}\\ 
\email{yifan.zhang@u.nus.edu}\texttt{,}
\email{shuaicheng.niu@ntu.edu.sg}\texttt{,}\\
\email{\{shuguangcui,lizhen\}@cuhk.edu.cn} 
}

\makeatletter
\renewcommand*{\@fnsymbol}[1]{\ensuremath{\ifcase#1\or *\or \dagger\or \ddagger\or
		\mathsection\or \mathparagraph\or \|\or **\or \dagger\dagger
		\or \ddagger\ddagger \else\@ctrerr\fi}}
\makeatother

\maketitle

\begin{abstract}
     Monocular 3D object detection (Mono 3Det) aims to identify 3D objects from a single RGB image. However, existing methods often assume training and test data follow the same distribution, which may not hold in real-world test scenarios. To address the out-of-distribution (OOD) problems, we explore a new adaptation paradigm for Mono 3Det, termed \textbf{Fully Test-time Adaptation}. It aims to adapt a well-trained model to unlabeled test data by handling potential data distribution shifts at test time without access to training data and test labels. However, applying this paradigm in Mono 3Det poses significant challenges due to OOD test data causing a remarkable decline in object detection scores. This decline conflicts with the pre-defined score thresholds of existing detection methods, leading to severe object omissions (\ie,~rare positive detections and many false negatives). Consequently, the limited positive detection and plenty of noisy predictions cause test-time adaptation to fail in Mono 3Det. To handle this problem, we propose a novel \textbf{Mono}cular \textbf{T}est-\textbf{T}ime \textbf{A}daptation (\textbf{MonoTTA}) method, based on two new strategies.
     1) Reliability-driven adaptation: we empirically find that \emph{high-score objects are still reliable} and the optimization of high-score objects can \emph{enhance confidence across all detections}. Thus, we devise a self-adaptive strategy to identify reliable objects for model adaptation, which discovers potential objects and alleviates omissions.
     2) Noise-guard adaptation: since high-score objects may be scarce, we develop a negative regularization term to exploit the numerous low-score objects via negative learning, preventing overfitting to noise and trivial solutions. Experimental results show that \ournet~brings significant performance gains for Mono 3Det models in OOD test scenarios, approximately 190\% gains by average on KITTI and 198\% gains on nuScenes.
     
\end{abstract}

\section{Introduction}
\label{sec:intro}

\begin{figure*}[t] 
  \includegraphics[width=\columnwidth]{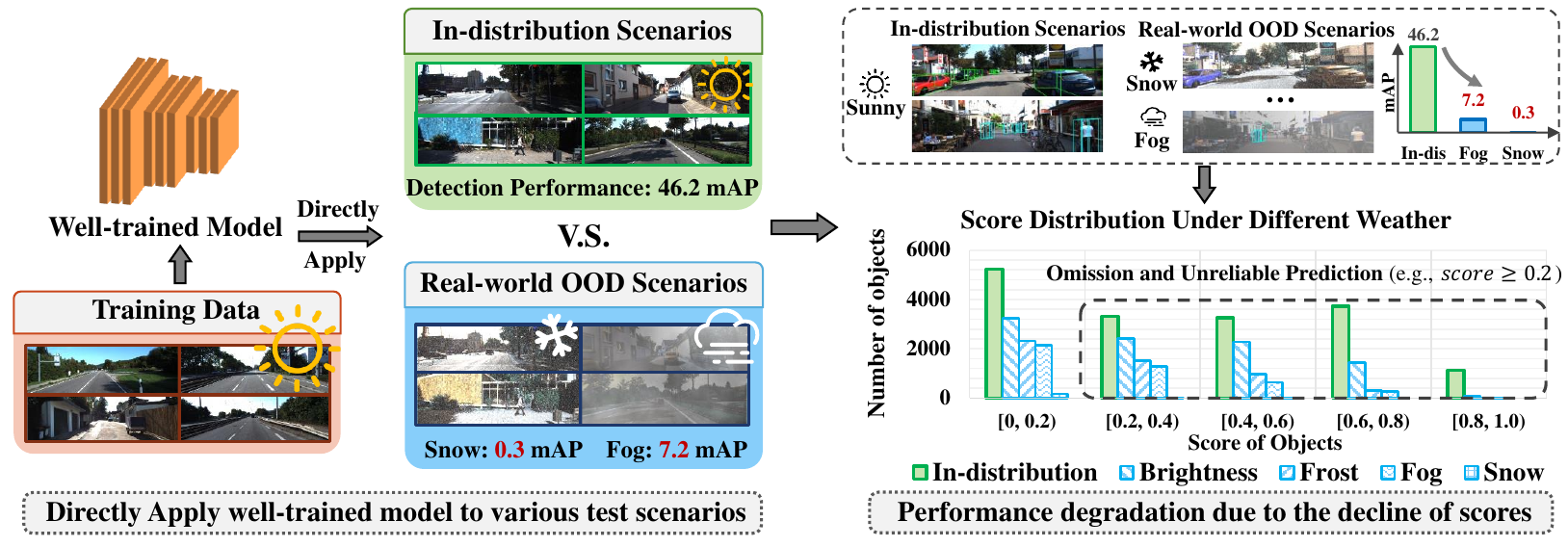} 
   \caption{
   An illustration of the generalizability issue of Mono 3Det models. Compared with in-distribution (In-dis) scenarios (\eg sunny), the detection scores within out-of-distribution (OOD) test data suffer severe degradation when the well-trained model (MonoFlex~\cite{zhang2021objects}) is directly applied to test scenarios affected by common natural disruptions, like weather changes (\eg snow and fog). Since existing Mono 3Det methods mainly adopt a pre-defined score threshold (\eg 0.2) for object detection, it leads to severe omissions and unreliable detections, thereby suffering serious performance degradation.
   Note that test images are the same but under different weather conditions.
   }
   \label{fig:our_setting}
   \vspace{-0.15in}
\end{figure*}

Three-dimensional (3D) Object Detection is a significant computer vision task, with the objective of identifying objects and determining their spatial and dimensional attributes through diverse sensor inputs~\cite{chen20153d,li2019stereo,wang2019pseudo,chen2023voxelnext,wu2023virtual}.
To reduce the cost of sensors, there is an increasing trend towards implementing autonomous driving systems via Monocular 3D Object Detection (Mono 3Det)~\cite{chen2016monocular,ye2022rope3d}, where only one single RGB image and the camera calibration information are given.
Even if this practical task is challenging, Mono 3Det methods have achieved promising results across various tasks and datasets~\cite{chen2020monopair,zhang2021objects,qin2022monoground,xu2023mononerd,luo2023latr}. Behind the success, a common presupposition is assuming that test images have the same distribution as the training images. However, this assumption could be possibly invalidated in many real-world scenarios due to prevalent natural corruptions such as weather changes, diminished sharpness, and other factors that introduce noise and contribute to uncalibrated cameras.
In such circumstances, the well-trained model often suffers substantial performance degradation as a consequence of the \emph{data distributional shifts} between the training images and the unlabeled test images. 
As shown in Fig.~\ref{fig:our_setting}, the model performance degrades from {46.2} mAP in in-distribution data to {0.3} mAP in Snow and {7.2} mAP in Fog.
Considering the widespread application of Mono 3Det in autonomous driving, the severe performance degradation due to out-of-distribution (OOD) test data may lead to unexpected traffic accidents and pose serious safety risks. Therefore, it is crucial to deal with the OOD generalization problem for Mono 3Det.

In addressing the OOD challenges specifically in test scenarios, one paradigm that has emerged as highly promising and gaining traction is \emph{Test-Time Adaptation} (TTA), which seeks to tackle data distribution shifts by adapting a well-trained model to unlabeled test images in real time~\cite{niu2024FOA}.
Test-time training (TTT)~\cite{sun2020test} represents an initial approach of TTA in classification tasks, by adjusting the well-trained model to predict rotations through additional model training, while its computation demands at the adaptation stage are prohibitive in Mono 3Det applications, particularly in autonomous driving. To enhance efficiency, Tent~\cite{wang2021tent} and EATA~\cite{niu2022efficient} have been developed for \emph{Fully Test-Time Adaptation} (Fully TTA) where only unlabeled test images and a well-trained model are provided. Besides, Ev-TTA~\cite{kim2022ev} and SOD\cite{veksler2023test} devise TTA methods to handle the event-based object recognition and weakly supervised salient object detection, respectively.
Considering the constraints on time of Mono 3Det, we explore the fully TTA paradigm which seeks to deal with OOD test data in real time.




To investigate this paradigm for Mono 3Det, we dig into the detection outcomes for objects within test scenarios with variations or corruptions that are commonly caused by weather or cameras. 
Specifically, we directly apply the well-trained model to the validation set of  KITTI which has been artificially injected with four distinct types of weather-related corruptions, namely Brightness, Frost, Fog, and Snow.
Subsequently, we plot their distributions of detection scores (c.f. Fig.~\ref{fig:our_setting}).
It is observed that the detection scores of test objects tend to \emph{markedly decline} as well as the \emph{high-score objects are scarce} in the extreme scenario (Snow) when the well-trained model is directly applied to the scenarios with corruptions.
This phenomenon indicates that: 
1) The pre-trained Mono 3Det model struggles to discriminate between objects and the background within OOD test data, presenting as quantities of \emph{omissions} and \emph{unconfident detections}.
2) Directly applying existing fully TTA methods to Mono 3Det could only get suboptimal performance since they struggle to optimize the model \emph{without enough high-score (positive) detections}, especially in certain extreme scenarios.

To handle it in Mono 3Det, we propose a \textbf{Mono}cular \textbf{T}est-\textbf{T}ime \textbf{A}daptation (\textbf{MonoTTA}) method, consisting of the reliability-driven adaptation and noise-guard adaptation strategies:
1) Reliability-driven adaptation.
Specifically, data distribution shifts lead to omissions and noisy detections while our empirical analysis suggests that \emph{high detection score objects are still reliable} (c.f.~Fig.~\ref{fig:our_motivation}~\textbf{(a)}). Moreover, even if we only optimize the model via high-score objects (\eg $\geq$0.5), both the numbers of low-score and high-score objects increase (c.f.~Fig.~\ref{fig:our_motivation}~\textbf{(b)}).
These investigations motivate us that exploiting high-score objects rather than all objects for model adaptation would be a more reliable way to alleviate data distribution shifts and discover potential objects.
Hence, we develop a self-adaptive strategy for the identification of reliable high-score objects in test images and devise the adaptive optimization loss $\mathcal{L}_{AO}$ to exploit the reliable subset for model adaptation, alleviating the detection score decline issue of OOD test data and digging out more potential objects.
2) Noise-guard adaptation.
In addition, data distribution shifts may also result in a scarcity of high-score objects, \ie,~the majority of objects presenting low scores as the `Snow' scenario in~Fig.~\ref{fig:our_motivation}~\textbf{(a)}. 
To this end, we develop a negative regularization term to make rational use of the numerous low-scoring objects in the Negative Learning manner~\cite{kim2019nlnl}. 
On the one hand, the negative regularization term $\mathcal{L}_{Nreg}$ allows the model to conduct adaptation via numerous noisy low-scoring objects. Thus, the model can achieve more high-score objects after alleviating distribution shifts. On the other hand, this term also prevents the model from overfitting to noise and trivial solutions, \ie,~assigning all classes of one object with high scores. 

We summarize the main contributions as follows:
\begin{itemize}
\item To the best of our knowledge, we are the first to explore Fully Test-Time Adaptation to address OOD generalization problems for Mono 3Det. We show that
the explored novel paradigm can bring significant improvements to Mono 3Det models in OOD test scenarios, \eg \textbf{137\%} and \textbf{244\%} average performance gains across 13 types of OOD shifts on KITTI.

\item Our empirical investigation reveals an important insight that high-score objects maintain their reliability amidst various corruptions, while optimizing these high-score objects significantly boosts model confidence across all detections. This motivates the first effective test-time adaptation method (\ie,~our MonoTTA) in Mono 3Det.

\item Extensive experiments on 13 types of corruptions of KITTI and 2 real scenarios (daytime $\leftrightarrow$ night) of nuScenes demonstrate the effectiveness of our \ournet~in boosting existing Mono 3Det methods~\cite{zhang2021objects,qin2022monoground} to handle test-time OOD problems. Even for instance-level methods~\cite{xu2023mononerd}, \ournet~also maintains sufficient improvement, which further confirms its applicability.

\end{itemize}

\section{Related Work}
\label{sec:related}

We first review the literature on Monocular 3D Object Detection, and then discuss Source-free Domain Adaptation and Test-Time Adaptation methods. 
More discussions on Unsupervised Domain Adaptation are put in Appendix A.

\noindent\textbf{Monocular 3D Object Detection}
aims to perceive 3D objects from a single 2D image. Existing Mono 3Det methods could be divided into two groups according to the use of extra information. On the one hand, some existing methods leverage extra pre-trained depth estimation modules~\cite{xu2018multi,ding2020learning,zou2021devil} to solve one of the most difficult problems in Mono 3Det, \ie,~depth estimation from a single image. Other methods utilize LiDAR information, \eg generating pseudo-LiDAR~\cite{wang2019pseudo,marethinking,reading2021categorical}. It is worth noting that Monoground~\cite{qin2022monoground} proposes to introduce the ground plane as prior information, and MonoNeRD~\cite{xu2023mononerd} proposes to utilize scene geometric clues to enhance the detector’s performance in the implicit reconstruction manner.
On the other hand, some Mono 3Det methods try to detect 3D objects without extra data. For example, SMOKE~\cite{liu2020smoke} proposes to detect 3D objects as the key points estimation task. Then, Monoflex~\cite{zhang2021objects} improves this idea by providing a flexible definition of object centers, which unifies the centers of regular and truncated objects. 
GrooMeD-NMS~\cite{kumar2021groomed} proposes a grouped mathematically differentiable Non-Maximal Suppression for Mono 3Det.

\noindent\textbf{Source-free Domain Adaptation (SFDA)}
aims to adapt the pre-trained source model to an unlabeled target domain without using the source data due to privacy issues~\cite{liang2020we,ijcai2021qiu}. 
SF-UDA$^3D$~\cite{saltori2020sf} first explores the SFDA framework to adapt the PointRCNN 3D detector to target domains, which consists of pseudo-labeling, reversible scale-transformations and motion coherency. Recently, the authors~\cite{hegde2023source} seek to exploit the source model more reliably and propose an uncertainty-aware teacher-student framework to filter incorrect pseudo labels during model adaptation, alleviating the negative impact of label noise.

Nonetheless, SFDA assumes all target data to be known in advance and makes predictions after multiple epochs of optimization, which may not be viable for real-time applications due to computational or time constraints.

\noindent\textbf{Test-Time Adaptation (TTA)}
seeks to improve model performance on test data via model adaptation through test samples even if data shifts exist. 
Early TTA methods~\cite{sun2020test,liu2021ttt++} endeavor to conduct additional model optimization on training data by self-supervised objectives, and then adapt the well-trained model to the test data via self-supervised objectives. However, in Mono 3Det applications like autonomous driving, the computation demands of such methods are prohibitive.
To solve this, \emph{Fully Test-Time Adaptation} methods are developed to adapt the well-trained model, where only unlabeled test images are available. Specifically, certain methods~\cite{schneider2020improving,nado2020evaluating,niu2023towards} tackle data distribution shifts by adapting the batch normalization layer statistics, while others alleviate this issue either by the entropy minimization of test data~\cite{fleuret2021test,wang2021tent} or maximizing the prediction consistency of different augmentations~\cite{zhang2021test,zhang2022memo}.
As for object detection tasks, Ev-TTA~\cite{kim2022ev} and SOD~\cite{veksler2023test} try to handle the event-based object recognition and weakly supervised salient object detection offline, respectively.

However, existing fully TTA methods struggle to optimize the model and solve distribution shifts in Mono 3Det due to numerous false negative detections.
To the best of our knowledge, our \ournet~stands as the first fully TTA method that handles distribution shifts for Mono 3Det models in real time.

\section{Monocular Test-Time Adaptation}

\subsection{Problem Statement} 
\label{sec:pd}

Without loss of generality, we denote the pre-trained (or well-trained) model as $f_{\Theta_0}(\textbf{x})$, which is achieved via training on labeled training images $\{(\textbf{x}_i^s, \textbf{y}_i^s)\}_{i=1}^{N}$.
The training images follow the training distribution $P\left(\textbf{x}\right)$ (\ie,~$\textbf{x}^s \sim P\left(\textbf{x}\right)$). Here, $\Theta_0$ represents the parameters of the pre-trained model and $N$ is the number of training data.
During the training stage, the model is optimized to fit (or overfit) the training data. Then, at the test stage, the model will be able to perform well if the unlabeled test images $\mathcal{D}_t\small{=}\{\textbf{x}_i\}_{i=1}^{N_t}$ follows the identical data distribution, \ie,~$\textbf{x} \sim P\left(\textbf{x}\right)$ where $N_{t}$ is the total number of test images.
However, in real applications, it is possible for the pre-trained model to encounter Out-Of-Distribution (OOD) test samples due to prevalent natural corruptions, namely distribution shifts,
\ie,~$\textbf{x} \sim Q\left(\textbf{x}\right)$ and $P\left(\textbf{x}\right) \neq Q\left(\textbf{x}\right)$. 

To address this issue, fully test-time adaptation~\cite{wang2021tent} seeks to tackle distribution shifts by adapting the pre-trained model $f_{\Theta_0}(\textbf{x})$ to unlabeled test images $\{\textbf{x}_i\}_{i=1}^{N_t}$ in real time. To achieve this goal, existing methods typically endeavor 
to update the model through the minimization of unsupervised objectives defined on test samples by 
$\min_{\hat{\Theta}} \mathcal{L}(\textbf{x};\Theta)$, 
where $\textbf{x} \sim Q\left(\textbf{x}\right)$ and $\hat{\Theta} \subseteq \Theta$. Here, $\hat{\Theta}$ denotes the subset of model parameters that should be updated (\ie,~\emph{batch normalization layers} following existing methods~\cite{wang2021tent,niu2022efficient}).
Most existing fully TTA methods focus on classification tasks, heavily relying on sufficient positive predictions for model adaptation.
Nevertheless, there is a significant difference between conventional classification tasks and Mono 3Det. As previously indicated, the detection scores of test images $\textbf{x}$ derived from $f_{\Theta_0}(\textbf{x})$ are prone to markedly decrease in the presence of corruptions as shown in Fig.~\ref{fig:our_setting}, leading to severe omissions (numerous false negatives) in Mono 3Det.
In such circumstances, the scarcity of positive detections presents a significant challenge for model adaptation to test distributions while adapting the model with unreliable low-score detections may significantly introduce the noise. 
Therefore, existing fully TTA methods tend to fail in the OOD generalization problems of Mono 3Det.

\begin{figure*}[t] 
  \centering
  \includegraphics[width=0.95\linewidth]{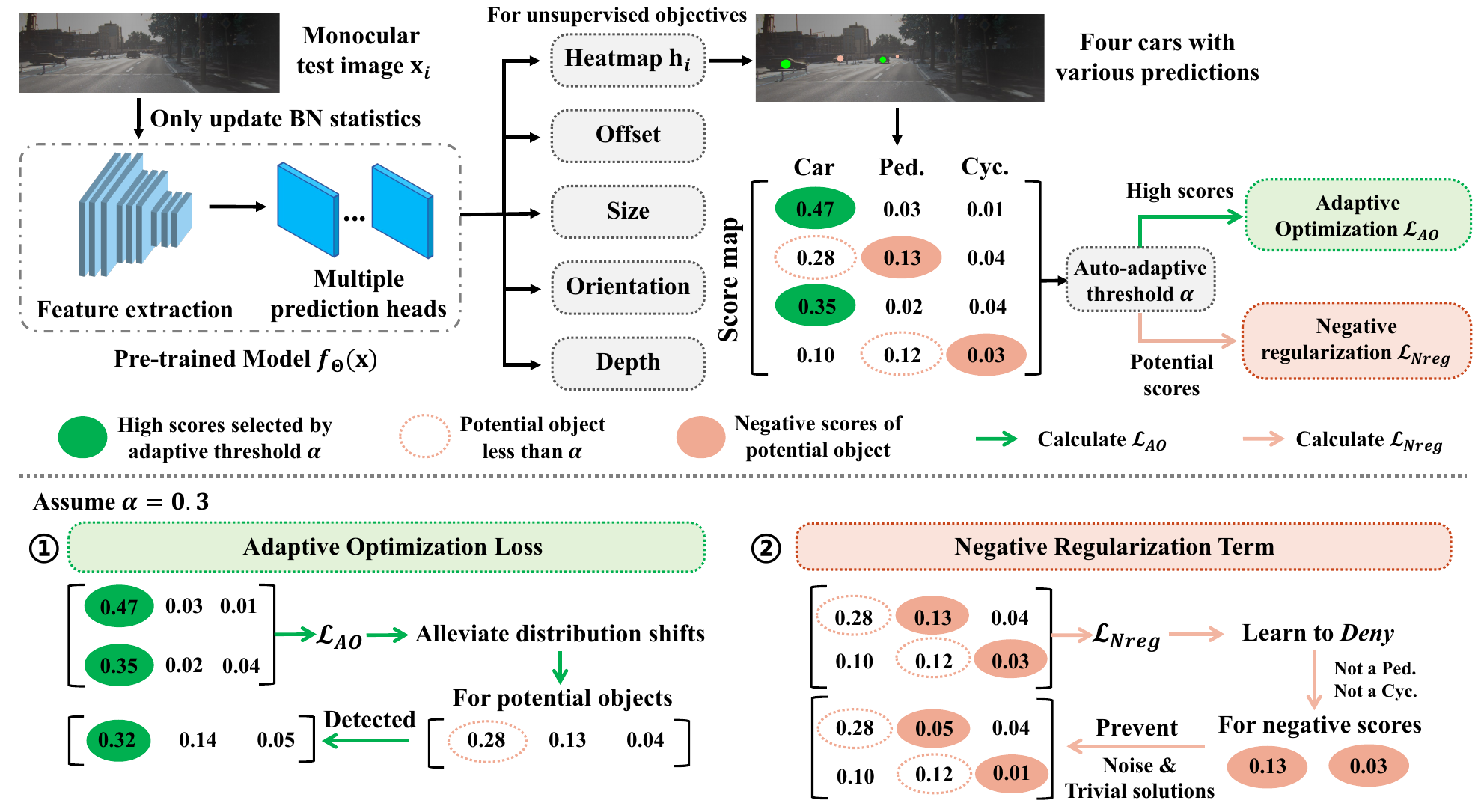} 
   \caption{
   An illustration of our \ournet. During the test phase, only the pre-trained model $f_{\Theta_0}(\textbf{x})$ and unlabeled test images $\{\textbf{x}_i\}_{i=1}^{N_t}$ are given.
   To conduct model adaptation, we initialize the model $f_{\Theta}(\textbf{x})$ by $\Theta_0$ and \emph{only update the parameters of batch normalization layers}. When a batch of test images arrives, we first compute test object scores and refine the adaptive threshold $\alpha$ to select the reliable high-score objects, thereby optimizing $\Theta$ via the adaptive optimization loss $\mathcal{L}_{AO}$. Meanwhile, we devise a negative regularization term $\mathcal{L}_{Nreg}$ to facilitate the model to avoid overfitting to noise and trivial solutions. Here, Ped. and Cyc. represent Pedestrian and Cyclist in KITTI.
   }
   \label{fig:our_method}
   \vspace{-0.1in}
\end{figure*}

\subsection{Overall Scheme}
After thoroughly examining the characteristics and challenges of Mono 3Det, we introduce a {Mono}cular {T}est-{T}ime {A}daptation (\ournet) method to address the OOD problems for Mono 3Det models, which seeks to solve the object score declining issue within unlabeled OOD test data.
As shown in Fig.~\ref{fig:our_method}, \ournet~consists of two strategies: 1) Reliability-driven adaptation and 2) Noise-guard adaptation. We first briefly introduce the two strategies below.

First, we develop a reliability-driven adaptation strategy (c.f.~Section~\ref{sec:conf}) to conduct reliable model adaptation for OOD test data based on dependable test objects. Our empirical investigations inspire us to exploit those relatively reliable test objects for alleviating distribution shifts, thereby discovering more potential objects. To this end, \ournet~excludes unreliable test objects involving an adaptive threshold $\alpha$ for any unlabeled test data. Subsequently, the model is optimized by the adaptive optimization loss $\mathcal{L}_{AO}$ via the selected reliable objects.

Second, we tend to utilize plenty of low-score objects to adapt the model in an indirect manner instead of directly optimizing the model since low-score objects are noisy. Hence, we devise a noise-guard adaptation strategy (c.f.~Section~\ref{sec:reg}) to prevent the model from overfitting to noisy predictions and falling into trivial solutions. Specifically, we randomly choose one of the negative classes of low-score objects and minimize the scores (\eg score 0.03 of $[0.10, 0.12, 0.03]$ c.f. Fig.~\ref{fig:our_method}) after simply filtering out extremely low-score detections. Even though the positive class is noisy (\ie,~score 0.12), this term is capable of optimizing the model indirectly, \ie,~learn to deny the negative category of the object.

Overall, the training scheme of \ournet~is as follows: 
\begin{equation}
\label{loss:total}
\min_{\hat{\Theta}} \mathcal{L}_{AO}(\hat{\Theta}) + \lambda \mathcal{L}_{Nreg}(\hat{\Theta}),
\end{equation}
where $\lambda$ is the balance hyper-parameter. The pseudo-code of \ournet~is summarized in Algorithm~\ref{al:training}.

\begin{figure}[t] 
  \centering
  \includegraphics[width=0.95\linewidth]{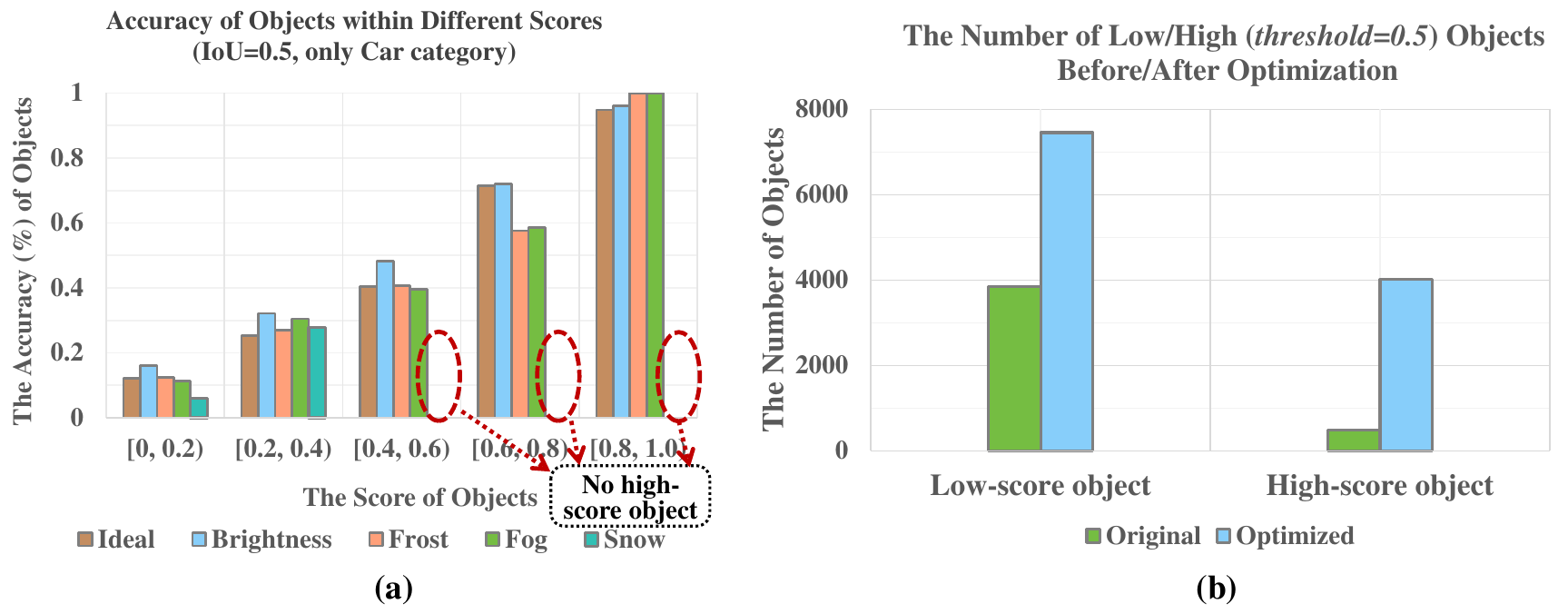}
  \vspace{-0.1in}
   \caption{
   Based on MonoGround~\cite{qin2022monoground}, we conduct two empirical studies (Car, KITTI), with the 3D IoU threshold of 0.5. 
   (a) We visualize the accuracy of the objects across varied scoring ranges, which shows that the accuracy of objects with high scores remains relatively stable even in the presence of diverse corruptions (Ideal means in-distribution scenarios).
   (b) We visualize the number of low \& high-score objects before and after optimization. Although only high-score objects are optimized, the model treats low-score objects with more confidence.
   }
   \label{fig:our_motivation}
   \vspace{-0.1in}
\end{figure}

\subsection{Reliability-Driven Adaptation}
\label{sec:conf}
To identify dependable test objects and conduct test-time model adaptation, we propose a reliability-driven adaptation strategy that consists of two components:
1) Reliable object identification and 2) Adaptive model optimization.

\noindent\textbf{Reliable object identification.}
When unlabeled test images arrive, it is difficult for the pre-trained model to get accurate detections on OOD test data due to the decline in detection scores.
To resolve this, we dig into the detection accuracy of the pre-trained model in test scenarios with various corruptions. Specifically, based on MonoGround~\cite{qin2022monoground}, we visualize the detection precision for the Car category within the KITTI dataset across varied scoring ranges, with the 3D Intersection over Union (IoU) threshold of 0.5.
As shown in Fig.~\ref{fig:our_motivation}~\textbf{(a)}, we find that high-score objects are more reliable and relatively stable even in the presence of diverse corruptions. Following this, we propose to select reliable high-score objects to conduct model adaptation via an adaptive threshold $\alpha$.
Specifically, we exploit the model $f_{\Theta}({\cdot})$ (initialized by $\Theta_0$) to infer a batch of test images $\{\textbf{x}_b\}_{b=1}^{B}$ and obtain the heatmap $\textbf{h}_i$ of each image $\textbf{x}_i$ by $\textbf{h}_{i}=f_{\Theta}(\textbf{x}_i)$, 
where $i$ ranges from $0$ to $B-1$ and $B$ denotes the batch size. Then, we could achieve object score maps $\textbf{s}_i \in \mathbb{R}^{B\times N_m}$ like the peaks in $\textbf{h}_i$~\cite{duan2019centernet} after normalization, where $K$ and $N_m$ denotes the number of classes and the maximum of detected objects, respectively.  
With the score map $\textbf{s}_i$, we update the adaptive threshold $\alpha_t$ at the iteration $t$ in the exponential moving average manner by:
\begin{align}\label{eq:cal_alpha}
\alpha_t=\left\{\begin{array}{ll}
\gamma, & \text { if } t=1 \\
\beta\bar{m}_t + (1-\beta)\alpha_{t-1}, & \text { if } t>1
\end{array},\right.
\end{align}
\begin{equation}
\label{eq:cal_mt}
\bar{m}_t=\frac{1}{B}\sum_{i=1}^{B}\frac{\sum_{j=1}^{N_m}{s}_{ij}\cdot\mathbb{I}({s}_{ij}\geq\gamma)}{\sum_{j=1}^{N_m}\mathbb{I}({s}_{ij}\geq\gamma)}.
\end{equation}
In Eqn.~(\ref{eq:cal_alpha}), $\bar{m}_t$ denotes the average score of all detected objects in a single batch of $B$ test samples at iteration $t$, while $\beta \in [0, 1]$ is a decay coefficient.
As for Eqn.~(\ref{eq:cal_mt}), ${s}_{ij} \small{\in}(0, 1)$ denotes the score of the $j$-th object in the $i$-th test image while $\mathbb{I}(\cdot)$ is the indicator function.
Note that $\gamma{\in}\mathbb{R}^1$ is a pre-defined object detection threshold adopted from existing methods at their original inference stage as well as $N_m$.

\noindent\textbf{Adaptive model optimization.}
As shown in Fig.~\ref{fig:our_motivation}~\textbf{(b)}, the optimization of high-score objects can also enhance the confidence of the model for relatively low-score objects. It motivates us that exploiting high-score objects rather than all objects for model adaptation would be a more reliable way to learn from OOD test data.
Therefore, with the adaptive threshold $\alpha_t$, we select the reliable subset of high scores from $\textbf{s}_i$ and calculate the adaptive optimization loss $\mathcal{L}_{AO}$ to adapt the model by:
\begin{equation}
\label{eq:loss_conf}
\mathcal{L}_{AO} = -\frac{1}{B}\sum_{i=1}^{B}\sum_{j=1}^{N_m}\log({s}_{ij}\cdot\mathbb{I}({s}_{ij}\geq\alpha_t)), 
\end{equation}
The adaptive optimization loss $\mathcal{L}_{AO}$ alleviates the potential data distribution shifts in OOD test data by allowing the model to confidently identify high-score test objects. As we mentioned, it solves the score decline issue of OOD test data in a more reliable way rather than directly optimizing all test objects, thereby avoiding overfitting to noise and discovering more potential test objects.

\begin{algorithm}[t]
    \caption{The pipeline of the proposed \ournet}\label{al:training}
    \begin{algorithmic}[1]
        \REQUIRE Unlabeled test data $\mathcal{D}_t\small{=}\{\textbf{x}_i\}_{i=1}^{N_t}$; Pre-trained model $f_{\Theta_0}(\textbf{x})$; Batch size $B$; Parameters $\lambda$, $\beta$, $\eta$.
    \FOR{a batch images $\{\textbf{x}_b\}_{b=1}^{B}$ in $\mathcal{D}_t$}
        \STATE Update the adaptive threshold $\alpha$ based on Eqn.~(\ref{eq:cal_alpha});
        \STATE Calculate adaptive optimization loss $\mathcal{L}_{AO}$ based on Eqn.~(\ref{eq:loss_conf});
        \STATE Calculate negative regularization term $\mathcal{L}_{Nreg}$ based on Eqn.~(\ref{eq:loss_reg});
        \STATE Update $\hat{\Theta}$ by optimizing Eqn.~(\ref{loss:total})
    \ENDFOR
    \RETURN Detection Results for all $\textbf{x} \in \mathcal{D}_t$.
     \end{algorithmic}
\end{algorithm}

\subsection{Noise-Guard Adaptation}
\label{sec:reg}
Through the optimization of adaptive optimization loss $\mathcal{L}_{AO}$, the model is refined to yield more confident detection outcomes. 
Nevertheless, high-scoring objects may be scarce due to the distribution shifts, 
making the adaptation procedure difficult. Meanwhile, exclusive reliance on $\mathcal{L}_{AO}$ for adaptation may result in trivial solutions, whereby the model indiscriminately assigns high scores to all categories.
Previous studies~\cite{kim2019nlnl,kim2021joint} indicate that deep neural networks could learn from noisy pseudo labels in classification tasks through negative learning. Thus, \ournet~proposes to learn from noisy low-score objects in a negative learning manner for Mono 3Det.
Specifically, we denote $\hat{\textbf{s}}_i\small{\in}\mathbb{R}^{B\times N_m\times K}$ as the multi-class score map for the test image $\textbf{x}_i$, \ie
the multi-class score map $\hat{\textbf{s}}_i$ contains not only the highest score of objects but also the scores for other classes. Here $\textbf{s}_{i}=\mathop{\arg\max}_{k}{\hat{\textbf{s}}_{ik}}$ where $k$ is the class index.
Employing a simple constant threshold $\eta$, we filter out extremely low-score objects and then randomly select a negative class $\bar{k}$ for each object.
Subsequently, we compute the regularization loss $e_k$ for each class $k$ with relatively low object scores $s_{ij} \in [\eta,\alpha_t)$ by:
\begin{equation}
e_k=-\sum_{i=1}^{B}\sum_{j=1}^{N_m}\bar{y}_{ij}\log(1- s_{ij\bar{k}} \cdot \mathbb{I}(\bar{k}=k)),  
\end{equation}
where $\bar{y}_{ij}\small{=}1-s_{ij\bar{k}}$ is a constant weight.
We further define $n_k$ as the frequency of negative scores corresponding to class $k$ in the test batch and balance $e_k$ by:
\begin{equation}
\label{eq:loss_reg}
\mathcal{L}_{Nreg} = \sum_{k=1}^{K} \frac{e_k}{n_k}.
\end{equation}

As we mentioned before, detections with low and intermediate scores tend to be more noisy (c.f. Fig.~\ref{fig:our_motivation}~\textbf{(a)}), which is unreliable for direct model adaptation. To this end, we introduce the regularization term $\mathcal{L}_{Nreg}$ to leverage noisy low-score detections and improve the model for assimilating potentially accurate information (c.f. Term 2 in Fig.~\ref{fig:our_method}). Moreover, this term also prevents the model from trivial solutions, \ie,~indiscriminately assigning high scores to all classes of a single object (\eg TENT~\cite{wang2021tent}).
In addition, high-score objects may be absent in certain extreme scenarios. For instance, the Snow scenario (c.f. Fig.~\ref{fig:our_motivation}~\textbf{(a)}) lacks objects with scores exceeding 0.4 when the pre-trained model is directly applied. Under such a circumstance, $\mathcal{L}_{Nreg}$ plays a more important role in model adaptation since it can alleviate distribution shifts even only low-score objects, \ie,~deny the negative category. In other words, $\mathcal{L}_{Nreg}$ enables the model to alleviate distribution shifts and achieve more relatively high-score objects, thereby laying a crucial foundation for $\mathcal{L}_{AO}$ in extremely challenging scenarios.

\section{Experiments}
\label{sec:exper}

\begin{table*}[t]
\setlength\tabcolsep{5pt}
\renewcommand\arraystretch{0.9}
    \begin{center}
    \caption{\label{tab:level_1} Comparison with baselines on the {KITTI-C validation} set, severity \textbf{level 1} regarding {Mean $AP_{3D|R_{40}}$}. The \textbf{bold} number indicates the best result.}
    \vspace{-0.05in}
    \scalebox{0.55}{
         \begin{tabular}{l|ccc|ccc|cccc|ccc|c}
         \toprule
         \multicolumn{15}{c}{\textbf{\textbf{Car}, IoU=0.5}} \\
         \midrule
         \multirow{2}{*}{Method}  &   
         \multicolumn{3}{c|}{Noise} & 
         \multicolumn{3}{c|}{Blur} & 
         \multicolumn{4}{c|}{Weather} & 
         \multicolumn{3}{c|}{Digital} & 
         \multirow{2}{*}{Avg.} \\
        \cmidrule(lr){2-4} \cmidrule(lr){5-7} \cmidrule(lr){8-11} \cmidrule(lr){12-14} 
        & Gauss. & Shot & Impul. & Defoc. & Glass & Motion & Snow & Frost & Fog & Brit. & Contr. & Pixel & Sat. \\
         \midrule
        Monoflex~\cite{zhang2021objects} & 3.84 & 7.48 & 5.31 & 2.59 & 3.73 & 11.05 & 0.23 & 7.77 & 7.57 & 24.87 & 6.92 & 28.16 & 31.46 & 10.84 \\
         ~~$\bullet~$ BN adaptation~\cite{schneider2020improving} & 13.58  & 21.93  & 18.78  & 15.87  & 8.59  & 24.32  & 5.42  & 21.45  & 24.63  & 31.80  & 30.58  & 41.04  & 30.71  & 22.21  \\
         ~~$\bullet~$ TENT~\cite{wang2021tent} & 17.80  & 27.09  & 23.18  & 21.66  & 11.90  & 28.75  & 6.84  & 26.58  & 30.78  & 35.65  & 34.72  & 41.71  & 35.91  & 26.35   \\
         ~~$\bullet~$ EATA~\cite{niu2022efficient} &  16.67  & 26.42  & 25.07  & 22.54  & 13.23  & 27.73  & 7.87  & 26.58  & 31.10  & 35.39  & 35.28  & 41.40  & 36.72  & 26.62  \\
         ~~$\bullet~$ \ournet~(Ours) &  \textbf{21.15}  & \textbf{28.65}  & \textbf{26.64}  & \textbf{25.91}  & \textbf{19.26}  & \textbf{31.48}  & \textbf{12.43}  & \textbf{30.24}  & \textbf{33.75}  & \textbf{36.84}  & \textbf{36.83}  & \textbf{41.97}  & \textbf{38.13}  & \textbf{29.48}  \\
        \midrule
        MonoGround~\cite{qin2022monoground} & 2.40 & 4.10 & 3.31 & 3.71 & 2.67 & 8.13 & 0.22 & 5.54 & 4.59 & 25.37 & 4.00 & 33.57 & 28.08 & 9.67   \\
         ~~$\bullet~$ BN adaptation~\cite{schneider2020improving} &  13.49 & 23.52 & 19.69 & 16.33 & 7.61 & 23.99 & 7.98 & 20.71 & 24.00 & 31.34 & 29.03 & 43.06 & 32.99 & 22.60  \\
         ~~$\bullet~$ TENT~\cite{wang2021tent} & 17.94  & 29.60  & 19.90  & 23.45  & 13.90  & 29.39  & 10.32  & 26.65  & 33.35  & 35.96  & 36.39  & 43.35  & 37.79  & 27.54  \\
         ~~$\bullet~$ EATA~\cite{niu2022efficient} & 16.03  & 26.08  & 18.08  & 20.28  & 12.40  & 27.37  & 9.22  & 23.79  & 29.49  & 33.65  & 32.58  & 43.61  & 36.00  & 25.28  \\
         ~~$\bullet~$ \ournet~(Ours) &  \textbf{26.13} & \textbf{33.11} & \textbf{28.60} & \textbf{30.38} & \textbf{25.48} & \textbf{32.44} & \textbf{18.72} & \textbf{32.60} & \textbf{37.75} & \textbf{37.87} & \textbf{39.57} & \textbf{43.67} & \textbf{37.98} & \textbf{32.64}  \\
        \midrule
        \multicolumn{15}{c}{\textbf{\textbf{Pedestrian}, IoU=0.25}} \\
         \midrule
         \multirow{2}{*}{Method}  &   
         \multicolumn{3}{c|}{Noise} & 
         \multicolumn{3}{c|}{Blur} & 
         \multicolumn{4}{c|}{Weather} & 
         \multicolumn{3}{c|}{Digital} & 
         \multirow{2}{*}{Avg.} \\
        \cmidrule(lr){2-4} \cmidrule(lr){5-7} \cmidrule(lr){8-11} \cmidrule(lr){12-14} 
        & Gauss. & Shot & Impul. & Defoc. & Glass & Motion & Snow & Frost & Fog & Brit. & Contr. & Pixel & Sat. \\
         \midrule
        Monoflex & 0.19 & 1.62 & 0.32 & 3.72 & \textbf{8.47} & 6.22 & 0.00 & 4.27 & 2.25 & 9.19 & 2.08 & 1.83 & 9.11 & 3.79  \\
         ~~$\bullet~$ BN adaptation~\cite{schneider2020improving} & 6.21 & 8.20 & 9.20 & 7.83 & 5.35 & 7.52 & 2.89 & 6.47 & 9.24 & 9.12 & 9.93 & 12.73 & 9.76 & 8.03  \\
         ~~$\bullet~$ TENT~\cite{wang2021tent} &  6.02 & 7.96 & 9.57 & 7.75 & 6.06 & 8.63 & 2.63 & 6.71 & 9.91 & 10.26 & \textbf{10.55} & 12.33 & 10.27 & 8.36   \\
         ~~$\bullet~$ EATA~\cite{niu2022efficient} &  6.05 & 7.96 & \textbf{9.74} & \textbf{7.93} & 6.06 & 8.01 & 2.48 & 6.24 & 9.94 & 9.70 & 10.02 & 12.41 & 10.12 & 8.21   \\
         ~~$\bullet~$ \ournet~(Ours) &  \textbf{6.54} & \textbf{8.41} & 9.39 & 7.63 & 7.12 & \textbf{8.99} &\textbf{3.25} & \textbf{7.64} & \textbf{10.26} & \textbf{10.55} & 10.06 & \textbf{13.28} & \textbf{10.66} & \textbf{8.75}  \\
        \midrule
        MonoGround & 0.61 & 0.93 & 0.73 & 8.41 & 7.57 & 8.03 & 0.00 & 3.19 & 1.39 & \textbf{13.82} & 1.83 & 3.70 & 7.26 & 4.42  \\
         ~~$\bullet~$ BN adaptation~\cite{schneider2020improving} & 5.09 & 7.08 & 7.77 & 7.63 & 6.63 & 9.45 & 2.24 & 6.72 & 8.40 & 9.72 & 11.06 & 16.70 & 12.43 & 8.53   \\
         ~~$\bullet~$ TENT~\cite{wang2021tent} &  7.27 & 10.10 & 10.02 & 8.53 & 8.30 & 11.03 & 3.54 & 8.73 & 9.11 & 11.74 & 12.12 & \textbf{17.70} & \textbf{15.03} & 10.25  \\
         ~~$\bullet~$ EATA~\cite{niu2022efficient} & 5.92 & 8.05 & 8.58 & 8.12 & 7.59 & 10.95 & 3.31 & 7.91 & 10.10 & 11.02 & 10.95 & 17.30 & 14.34 & 9.55   \\
         ~~$\bullet~$ \ournet~(Ours) & \textbf{8.58} & \textbf{11.18} & \textbf{11.79} & \textbf{9.22} & \textbf{9.40} & \textbf{13.20} & \textbf{4.83}& \textbf{9.95} & \textbf{14.46} & 12.85 & \textbf{13.25} & 17.13 & 14.85 & \textbf{11.59}   \\
        \midrule
        \multicolumn{15}{c}{\textbf{\textbf{Cyclist}, IoU=0.25}} \\
         \midrule
         \multirow{2}{*}{Method}  &   
         \multicolumn{3}{c|}{Noise} & 
         \multicolumn{3}{c|}{Blur} & 
         \multicolumn{4}{c|}{Weather} & 
         \multicolumn{3}{c|}{Digital} & 
         \multirow{2}{*}{Avg.} \\
        \cmidrule(lr){2-4} \cmidrule(lr){5-7} \cmidrule(lr){8-11} \cmidrule(lr){12-14} 
        & Gauss. & Shot & Impul. & Defoc. & Glass & Motion & Snow & Frost & Fog & Brit. & Contr. & Pixel & Sat. \\
         \midrule
        Monoflex & 0.28 & 1.64 & 0.47 & 0.59 & 4.97 & 3.60 & 0.00 & 7.42 & 3.81 & \textbf{13.07} & 3.79 & 3.80 & 8.39 & 3.99  \\
         ~~$\bullet~$ BN adaptation~\cite{schneider2020improving} & 2.39 & 6.26 & 4.36 & 5.78 & 6.76 & 9.09 & \textbf{1.70} & 8.53 & 9.16 & 12.91 & 11.26 & 10.55 & 11.02 & 7.67   \\
         ~~$\bullet~$ TENT~\cite{wang2021tent} &  2.72 & \textbf{7.94} & 5.63 & 6.27 & 7.20 & 9.49 & 1.07 & 8.94 & 10.96 & {12.75} & \textbf{12.72} & 9.64 & \textbf{11.28} & 8.20   \\
         ~~$\bullet~$ EATA~\cite{niu2022efficient} &   2.33 & 7.46 & 5.46 & \textbf{7.19} & \textbf{7.23} & 7.51 & 1.24 & 8.60 & 10.18 & 12.86 & 11.02 & 10.89 & 10.30 & 7.87   \\
         ~~$\bullet~$ \ournet~(Ours) &  \textbf{3.01} & 7.24 & \textbf{5.98} & 7.00 & 6.09 & \textbf{9.51} & 1.45 & \textbf{10.63} & \textbf{11.22} & 12.85 & 11.85 & \textbf{11.59} & 10.81 & \textbf{8.40}   \\
        \midrule
        MonoGround &  0.12 & 0.46 & 0.48 & 0.33 & 0.72 & 2.03 & 0.00 & 0.56 & 0.35 & 5.55 & 0.52 & 2.08 & 5.24 & 1.42  \\
         ~~$\bullet~$ BN adaptation~\cite{schneider2020improving} &  1.76 & 3.58 & 2.08 & 3.61 & 3.05 & 5.41 & 0.57 & 3.82 & 4.47 & 6.30 & 5.60 & 11.02 & 8.87 & 4.63   \\
         ~~$\bullet~$ TENT~\cite{wang2021tent} &  1.79 & 4.85 & 3.00 & 3.36 & 3.49 & 6.05 & 0.49 & 4.43 & 6.20 & 7.19 & 6.50 & 10.43 & 9.23 & 5.15   \\
         ~~$\bullet~$ EATA~\cite{niu2022efficient} &  1.89 & 4.20 & 2.41 & 4.13 & 3.17 & 5.73 & 0.38 & 4.06 & 6.27 & 6.41 & 6.14 & 10.93 & 7.72 & 4.88   \\
         ~~$\bullet~$ \ournet~(Ours) & \textbf{3.93} & \textbf{5.78} & \textbf{4.55} & \textbf{5.43} & \textbf{4.70} & \textbf{6.09} & \textbf{0.69} & \textbf{4.66} & \textbf{7.53} & \textbf{7.69} & \textbf{7.74} & \textbf{11.71} & \textbf{9.43} & \textbf{6.15}  \\        
         \bottomrule
         \end{tabular}
         }
    \end{center}
    \vspace{-0.25in}
\end{table*}

We conduct experiments based on KITTI~\cite{geiger2012we} and nuScenes~\cite{caesar2020nuscenes}. 
The results presented in this manuscript represent the \emph{average value across three difficulty levels}, \ie,~Easy, Moderate, Hard.
Note that we provide more results of higher severity levels of KITTI and more detailed results in Appendix \textbf{\emph{D}}.

\noindent\textbf{Datasets.}
For KITTI, we adopt the protocol established by Monoflex~\cite{zhang2021objects} to split the training images into the training set (3712) and validation set (3769) to perform model training and adaptation, respectively. Following the official KITTI evaluation criteria, we evaluate detection results on three levels of difficulty. Then, we construct the KITTI-C dataset through the incorporation of 13 distinct types of data corruptions~\cite{hendrycks2018benchmarking} to the validation set and each corruption has 5 severity levels. The model is trained on the original training set and then tested on the KITTI-C validation set within one of the corruptions.

As for nuScenes, we adopt the front-view images and construct the Daytime and Night scenarios via their scene descriptions following~\cite{liu2023bevfusion}.
Specifically, there exist 24.7k/5.4k train/test images in Daytime while 3.3k/0.6k train/test images in Night. Based on these splits, we construct two real-world adaptation tasks, \ie,~Daytime $\rightarrow$ Night and Night $\rightarrow$ Daytime. For simplification, we transfer the nuScenes dataset into the KITTI format and only consider the Car category.
More details of data construction are provided in Appendix \textbf{\emph{B}}.

\begin{figure*}[t] 
  \centering
  \includegraphics[width=\linewidth]{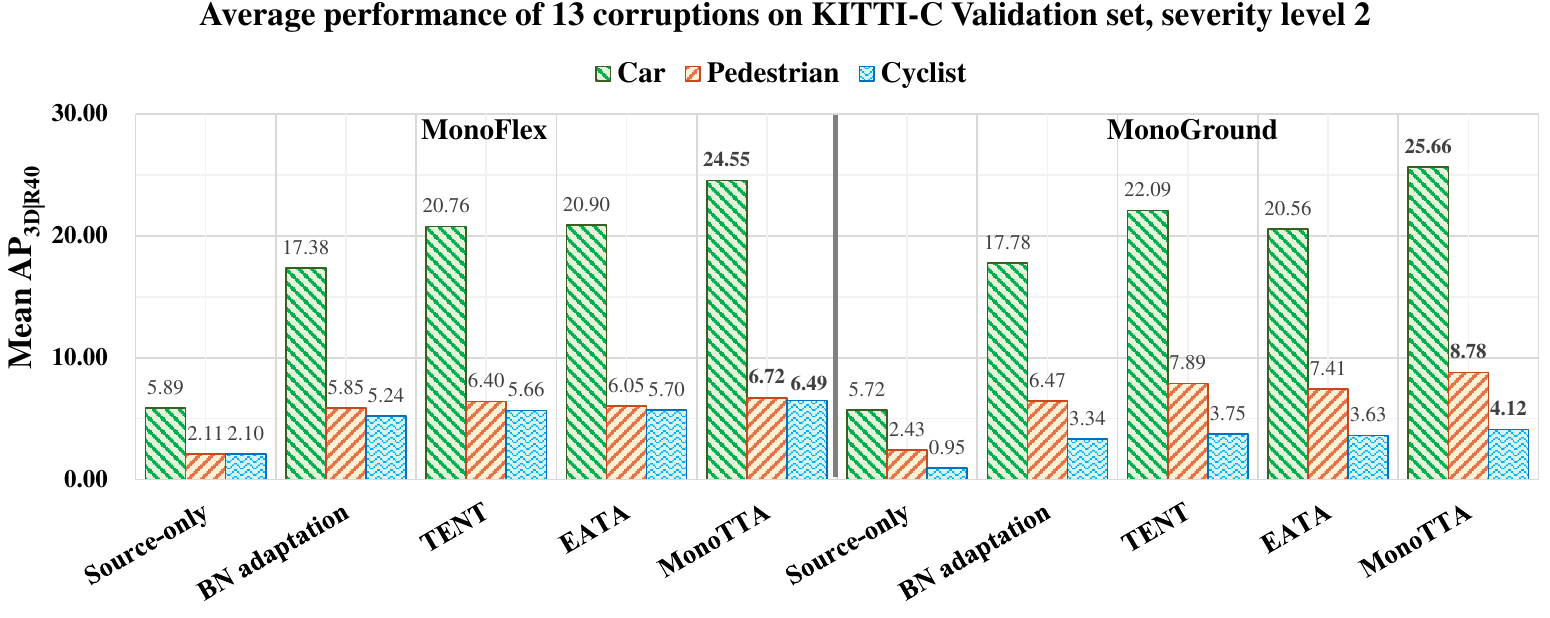} 
   \caption{
    We visualize the comparison with baselines on the {KITTI-C validation} set, severity \textbf{level 2} regarding Mean $AP_{3D|R_{40}}$. The \textbf{bold} number indicates the best result.
   }
   \label{fig:level_2}
   \vspace{-0.05in}
\end{figure*}

\noindent\textbf{Implementation details.}
We implement our method and other baselines in PyTorch~\cite{paszke2019pytorch}. In \ournet, we conduct model adaptation based on the public pre-trained weights and the parameter settings provided by their authors~\cite{zhang2021objects,qin2022monoground,xu2023mononerd}.
Besides, we employ the Stochastic Gradient Descent (SGD) optimizer with a half learning rate of the initial rate used in base training over different methods, a momentum of 0.9 and a batch size of 16 for KITTI, 4 for nuScenes. Parameters $\lambda$, $\beta$, $\eta$ are assigned default values of 1, 0.1, and 0.05, respectively.
More training details of \ournet~are provided in Appendix \textbf{\emph{C}}.

\noindent\textbf{Compared methods.} Based on three typical or state-of-the-art (SOTA) Mono 3Det methods~\cite{zhang2021objects,qin2022monoground,xu2023mononerd},
we fully compare \ournet~with following methods: 
1) source-only, \ie,~directly apply the pre-trained model to the test data within corruptions;
2) BN adaptation~\cite{schneider2020improving} updates batch normalization statistics via target data;
3) TENT~\cite{wang2021tent} minimizes the entropy loss of test data;
4) EATA~\cite{niu2022efficient} identifies reliable samples to update the model by entropy loss minimization, which is a SOTA method in Fully Test-Time Adaptation. Here we compare \ournet~with its variant {E}fficient {T}est-time {A}daptation.

\noindent\textbf{Evaluation protocols}.
In order to fully evaluate the proposed method, we report our experimental results in the Average Precision (AP) for 3D bounding boxes, denoted as $AP_{3D|R_{40}}$. As previously mentioned, the results present the mean values across three distinct levels of difficulty and the Intersection over Union (IoU) thresholds are set to 0.5, 0.25, 0.25 for Cars, Pedestrians, Cyclists.


\begin{figure*}[t] 
  \centering
  \includegraphics[width=0.9\linewidth]{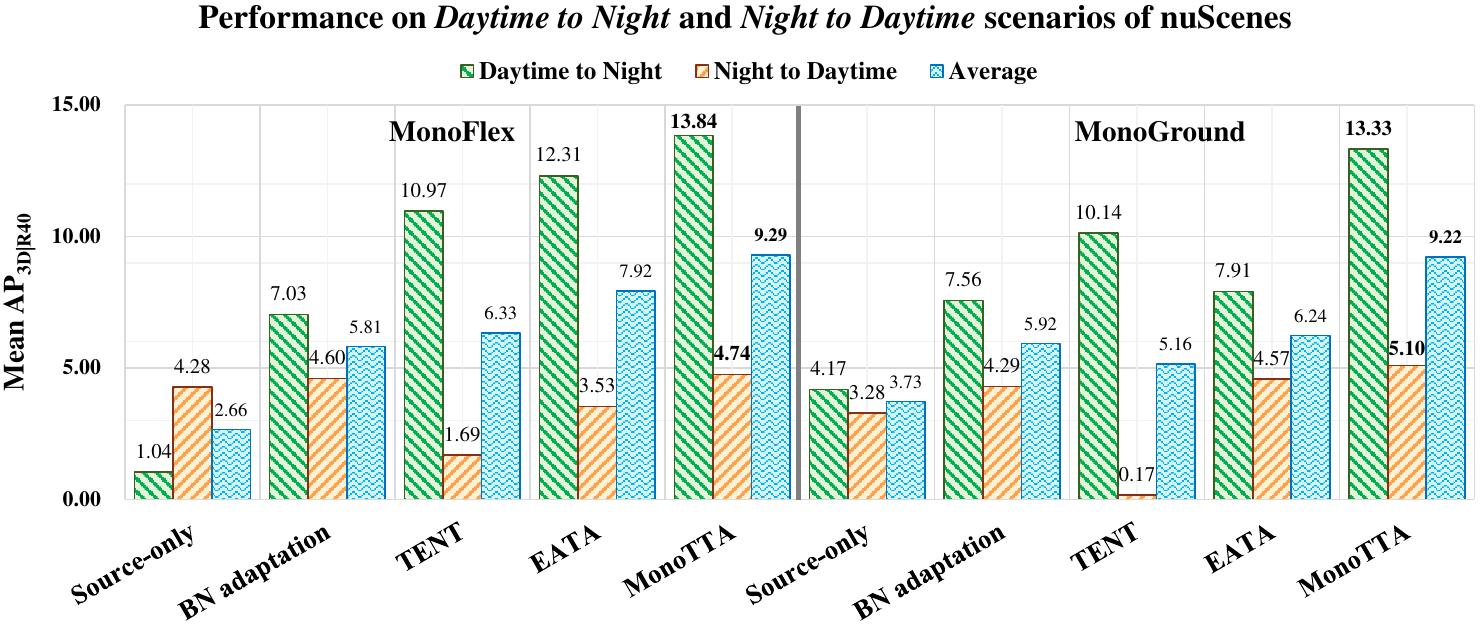} 
   \caption{
    Comparison with baselines on \textbf{D}aytime $\rightarrow$ \textbf{N}ight and \textbf{N}ight $\rightarrow$ \textbf{D}aytime of {nuScenes}, regarding {Mean $AP_{3D|R_{40}}$}. The \textbf{bold} number indicates the best result.
   }
   \label{fig:nus_kitti}
   \vspace{-0.05in}
\end{figure*}

\subsection{Comparisons with Previous Methods}
We first compare our \ournet~with previous methods in severity level 1 of KITTI-C.
The results are reported in Table~\ref{tab:level_1}, which gives the following observations:
1) Due to distribution shifts, directly applying the pre-trained model to the test data (\ie,~source-only) suffers severe performance degradation in all categories.
2) Existing TTA methods are able to mitigate the negative effect of distribution shifts for Mono 3Det to some degree. However, they only achieve suboptimal performance since they tend to increase the scores of all positive detections, containing severe noise.
3) \ournet~consistently outperforms all compared methods over all categories within various base models in terms of mean $AP_{3D|R_{40}}$. 
Specifically, \ournet~achieves the best or comparable performance in all categories under all corruptions, attaining a large performance gain over TENT and EATA (\eg improving an average $AP_{3D|R_{40}}$ about 5.1 and 7.4 of the Car category based on MonoGround).

\subsection{More Severe Corruption and Real Scenario}
On the one hand, to fully validate the effectiveness of our \ournet, we visualize experimental results under more severe corruption conditions at severity level 2 as shown in Fig.~\ref{fig:level_2}, which clearly gives additional observations: 
1) With the escalation of severity level, the pre-trained models suffer a larger performance decline within various corruptions, enlarging the difficulty of TTA.
2) The performance improvements of existing TTA methods become relatively limited, particularly in Pedestrian and Cyclist classes.
3) Even if the tasks are more challenging, \ournet~still stably obtains the best average performance within all corruptions since $\mathcal{L}_{Nreg}$ plays an important role in alleviating distribution shifts for certain extreme scenarios, \ie,~only low-score objects exist.

On the other hand, we further validate different methods within real scenarios as shown in Fig.~\ref{fig:nus_kitti}. The experimental results also give the following observations:
1) Under real corruptions, the pre-trained model still suffers severe performance degradation due to the data distribution shifts.
2) TENT tends to increase the confidence of all positive detections and thus overfits to noise, \ie,~failing to handle the extremely challenging task N$\rightarrow$D.
3) EATA still achieves sub-optimal performance while our \ournet~brings sufficient average performance improvement on both MonoFlex (6.23 mAP) and MonoGround (8.26 mAP), maintaining the best performance in real scenarios and 
further demonstrating the effectiveness and superiority of the proposed \ournet. More detailed results of this part are put in Appendix \textbf{\emph{D}}.

\begin{table}[t]
\setlength\tabcolsep{9pt}
\renewcommand\arraystretch{0.95}
    \begin{center}
    \caption{\label{tab:nerd_instance} Comparison with baselines based on the instance-level method (\ie,~batch size is 1) on the {KITTI-C validation} set, regarding {Mean $AP_{3D|R_{40}}$} and the severity {levels 1 and 2}. The \textbf{bold} number indicates the best result.}
        \scalebox{0.55}{
         \begin{tabular}{l|cccc|ccccccc}
         \toprule
         \multirow{2}{*}{Method} & 
         \multicolumn{4}{c|}{Level 1} &
         \multicolumn{4}{c}{Level 2}
         \\
         \cmidrule(lr){2-5} \cmidrule(lr){6-9} 
         & Car & Pedes. & Cyclist & Avg. & Car & Pedes. & Cyclist & Avg. \\
         \midrule
         MonoNeRD & 19.84  & 5.96  & 2.27  & 9.36  & 13.02  & 3.83  & 1.61  & 6.15 \\
         ~~$\bullet~$ BN adaptation~\cite{schneider2020improving} & 30.73  & 8.85  & 3.81  & 14.46  & 26.47  & 6.94  & 2.91  & 12.11  \\
         ~~$\bullet~$ TENT~\cite{wang2021tent} &  35.72  & 9.99  & \textbf{4.75}  & 16.82  & 31.85  & 7.81  & \textbf{3.40}  & 14.35    \\
         ~~$\bullet~$ EATA~\cite{niu2022efficient} &   34.60  & 10.04  & 4.20  & 16.28  & 30.66  & 7.86  & 3.28  & 13.93   \\
         ~~$\bullet~$ \ournet~(Ours) & \textbf{37.40}  & \textbf{10.39}  & 4.35  & \textbf{17.38}  & \textbf{33.99}  & \textbf{8.25}  & 3.33  & \textbf{15.19}   \\
         \bottomrule
         \end{tabular}
         }
    \end{center}
    \vspace{-0.1in}
\end{table}

\subsection{Application to Instance-Level Inference Method}
In this section, we seek to investigate whether the proposed \ournet~can be used to effectively enhance Mono 3Det methods which only process a single image sequentially, \ie,~$B=1$. 
To be specific, it may be crucial for Mono 3Det methods to make immediate decisions based on the most recent scene (image) in real-world scenarios like autonomous driving. To this end, it is essential for TTA methods devised for Mono 3Det to allow a single image as input and then conduct model adaptation. 
To validate it, we integrate \ournet~into the SOTA Mono 3Det method namely MonoNeRD~\cite{xu2023mononerd} which accepts one image each time at the test phase. As shown in Table~\ref{tab:nerd_instance}, \ournet~achieves the best or comparable performance across all categories at both severity levels 1 and 2,  illustrating the applicability of our method to boost these approaches for handling OOD test data. Detailed results can be also found in Appendix \textbf{\emph{D}}.

\begin{table}[t]
\setlength\tabcolsep{8pt}
\renewcommand\arraystretch{0.95}
    \begin{center}
    \caption{\label{tab:ablation}
    Based on MonoGround~\cite{qin2022monoground}, we conduct ablation studies of $\mathcal{L}_{AO}$ and $\mathcal{L}_{Nreg}$ on the {KITTI-C validation} set, regarding {$AP_{3D|R_{40}}$}.
    }
    \scalebox{0.55}{
         \begin{tabular}{ccc|cccc|ccccccc}
         \toprule
         \multirow{2}{*}{Backbone} & \multirow{2}{*}{$\mathcal{L}_{AO}$} & \multirow{2}{*}{$\mathcal{L}_{Nreg}$} & 
         \multicolumn{4}{c|}{Level 1} &
         \multicolumn{4}{c}{Level 2}
         \\
         \cmidrule(lr){4-7} \cmidrule(lr){8-11} 
         & & & Car & Pedes. & Cyclist & Avg. & Car & Pedes. & Cyclist & Avg. \\
         \midrule
         \cmark &  &  & 9.67  & 4.42  & 1.42  & 5.17  & 5.72  & 2.43  & 0.95  & 3.03    \\
         \cmark & \cmark &  & 27.94  & 10.17  & 5.10  & 14.40  & 21.27  & 7.79  & 3.82  & 10.96   \\
         \cmark &  & \cmark &  23.26  & 8.60  & 4.60  & 12.15  & 18.93  & 6.81  & 3.55  & 9.77    \\
         \cmark & \cmark & \cmark & \textbf{32.64}  & \textbf{11.59}  & \textbf{6.15}  & \textbf{16.79}  & \textbf{25.66}  & \textbf{8.78}  & \textbf{4.12}  & \textbf{12.85}   \\
         \bottomrule
         \end{tabular}
         }
    \end{center}
\end{table}

\begin{figure*}[t] 
  \centering
  \includegraphics[width=0.9\linewidth]{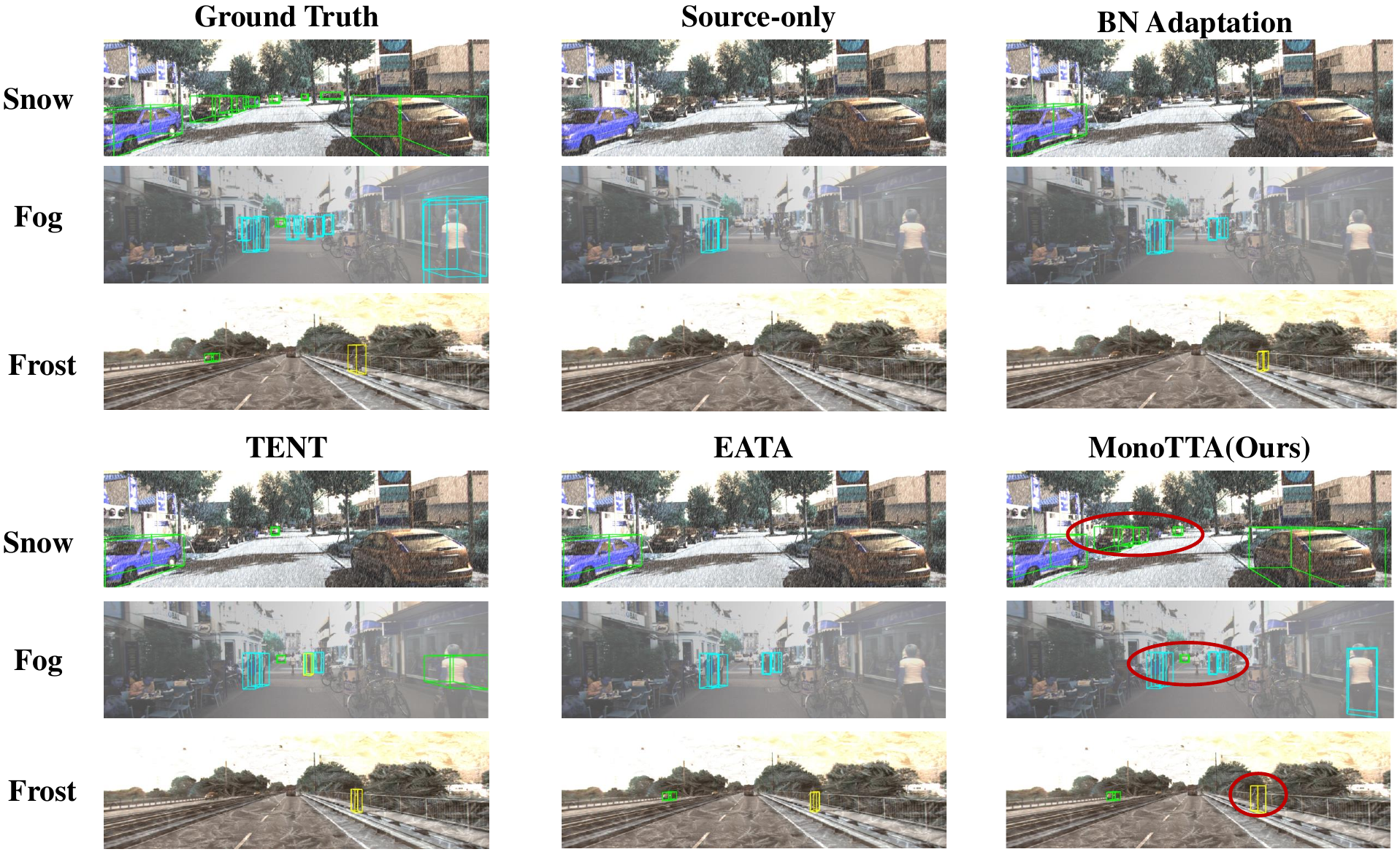} 
   \caption{
    Qualitative results of baselines and the proposed \ournet~based on MonoGround~\cite{qin2022monoground}.
    We visualize the results on KITTI-C validation set, where predicted cars, pedestrians and cyclists are in lime green, sky blue and yellow, respectively.
   }
   \label{fig:visualization}
   \vspace{-0.15in}
\end{figure*}

\subsection{Ablation Studies and Quantitative Results}
\noindent{\textbf{Ablation studies.}}
To examine the effectiveness of the losses in \ournet, we show the results of the models optimized by different losses.
As shown in Table~\ref{tab:ablation}, introducing $\mathcal{L}_{AO}$ or $\mathcal{L}_{Nreg}$ enhances the model performance compared to directly applying the pre-trained model (\ie,~source only).
On the one hand, such a result verifies that our strategy is able to alleviate the score decline issue in test OOD data with high-score objects. 
On the other hand, introducing $\mathcal{L}_{Nreg}$ could obtain a higher average $AP_{3D|R_{40}}$ value, which also verifies that the negative regularization term is able to enhance the pre-trained model even only with low-score objects.
When combining the losses (\ie{$\mathcal{L}_{AO}$, $\mathcal{L}_{Nreg}$}) together, we obtain the best performance.

\noindent{\textbf{Quantitative Results.}}
We provide visualizations within Snow, Fog and Frost based on Monoground as shown in Fig.~\ref{fig:visualization}. 
It is evident that the source-only setting suffers severe omissions, while BN adaptation, TENT and EATA alleviate distribution shifts to some degree and give more detections.
As for our \ournet, it can produce superior detections even in severe conditions, including fewer omissions and accurate detections as highlighted by red circles.

\section{Conclusion}
In this paper, we propose a {mono}cular {t}est-{t}ime {a}daptation method to improve the pre-trained model on the shifted test data for monocular 3D object detection. Specifically, our method consists of two strategies: 1) Reliability-driven adaptation. To discover more potential objects, we devise a self-adaptive strategy to identify reliable objects for adaptive model adaptation. 2) Noise-guard adaptation. To avoid overfitting to noise and trivial solutions, we devise the negative regularization term to mitigate the negative effects of noisy detections and alleviate distribution shifts.
Experiments on KITTI-C and nuScenes datasets demonstrate the effectiveness of \ournet~in handling fully test-time adaptation for monocular 3D object detection. 

\noindent\textbf{Future directions.} 
1) Our work focuses on 2D images, while future studies could explore 3D information in handling distribution shifts.
2) This work explores TTA by assuming one OOD distribution at a time, where the forgetting issue is not severe as the source model weights are recoverable. Exploring scenarios with dynamically OOD distributions offers a compelling future direction, where the forgetting issue would become more severe. 
%
%
\bibliographystyle{splncs04}
\bibliography{main}

\title{Supplementary Material} 
\titlerunning{Abbreviated paper title}



\clearpage

\author{
}
%
\institute{
}

\maketitle
\renewcommand{\thesection}{\Alph{section}}

In the supplementary, we first provide more related work and discussions to clarify existing solutions to data distribution shifts related to 3D object detection.
In addition, we provide more experimental details, visualizations, and results of \ournet. We organize our supplementary materials as follows.

\begin{itemize}
    \item In Section~\ref{supp:sec:relations}, we review Unsupervised Domain Adaptation in 3D Object Detection and provide more discussions.
    \item In Section~\ref{supp:sec:more_datasets}, we provide more details of the construction of the KITTI-C and two tasks of nuScenes dataset.
    \item In Section~\ref{supp:sec:more_impl},we provide more experimental details of our proposed \ournet
    \item In Section~\ref{supp:sec:more_res}, we show more experimental results to demonstrate the effectiveness of the proposed \ournet.
\end{itemize}

\section{More Related Work and Discussions}
\label{supp:sec:relations}

In this section, we review the literature on Unsupervised Domain Adaptation in 3D Object Detection to provide more discussions to clarify existing solutions to data distribution shifts related to 3D objects.

\noindent\textbf{Unsupervised Domain Adaptation (UDA)}
seeks to improve the source model performance on the unlabeled target domain based on a label-rich relevant source domain by alleviating the domain shifts~\cite{lin2022prototype,Zhang2020CollaborativeUD}. 
Specifically, ST3D~\cite{yang2021st3d} has been introduced to conduct UDA on 3D object detection from point clouds via alternatively pseudo labels updating and model training with curriculum data augmentation. Besides, DG-BEV~\cite{wang2023towards} decouples the depth estimation and performs dynamic perspective augmentation to facilitate domain generalization within multi-view 3D object detection tasks. Additionally, Spectral UDA~\cite{zhang2022spectral} has been advanced as a UDA technique that adapts in the spectral space, seeking to generalize across domains even in 3D object detection. 
As for UDA in Mono3D, STMono3D~\cite{liunsupervised} first proposes a self-teaching framework to address the depth-shift issue caused by the geometric misalignment of domains.

However, UDA relies on the achievement of the source training data to overcome data distribution shifts while the source data may be unavailable in practice due to privacy issues.
In addition, UDA assumes unlabeled target data and test data following the same data distribution, which may not be held in real.

\begin{figure*}[t] 
  \centering
  \includegraphics[width=1.0\linewidth]{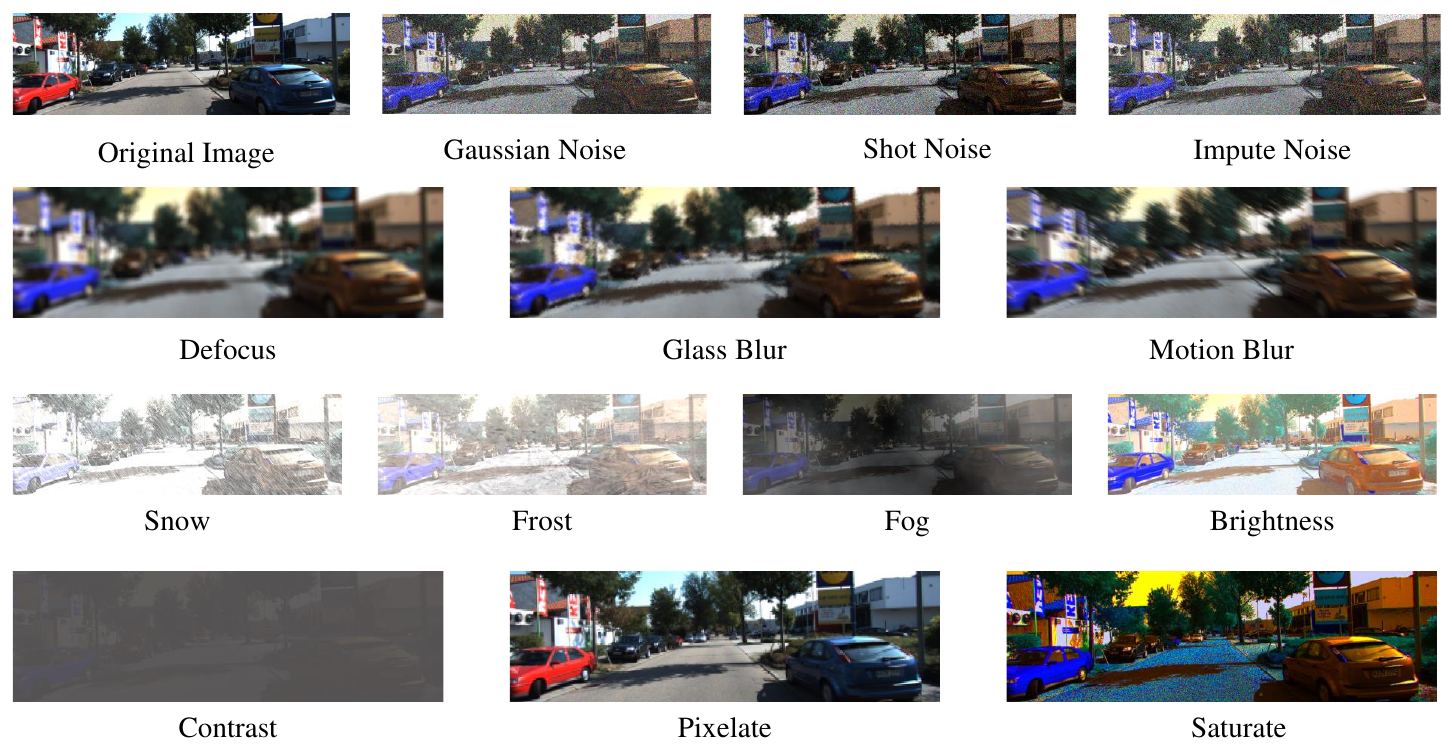} 
   \caption{
   An illustration of 13 distinct types of corruptions in the severity level 5 of the KITTI-C dataset.
   }
   \label{supp:fig:level5}
   \vspace{0.1in}
\end{figure*}

\begin{figure*}[t] 
  \centering
  \includegraphics[width=1.0\linewidth]{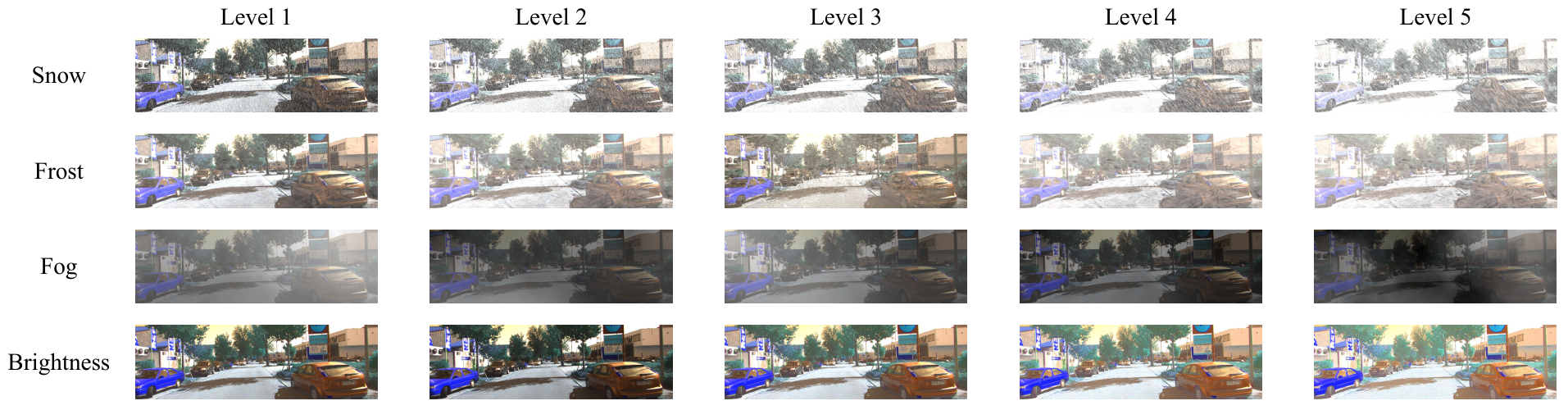} 
   \caption{
   An illustration of 4 common types of corruptions in real applications from 1 to 5 severity levels of the KITTI-C dataset.
   }
   \label{supp:fig:level1-5}
\end{figure*}

\section{More Details on Dataset Construction}\label{supp:sec:more_datasets}
In this section, we first provide more visualizations of the KITTI-C dataset to illustrate the construction details.  Then, we offer more details of 2 real scenarios (daytime $\leftrightarrow$ night) of the nuScenes dataset.

For the KITTI-C dataset, as shown in Figure~\ref{supp:fig:level5}, we construct the KITTI-C dataset through the integration of 13 distinct categories of data corruptions~\cite{hendrycks2018benchmarking}.
On the other hand, we also provide the visualizations of several frequent corruptions at different severity levels (from level 1 to level 5), including Snow, Frost, Fog and Brightness as shown in Figure~\ref{supp:fig:level1-5}.
Based on a comprehensive set of 65 practical scenarios (\ie 13 corruptions $\times$ 5 severity levels), we are able to fully verify the effectiveness of each method in addressing dataset distribution shifts encountered in the Fully Test-time Adaptation task for Monocular 3D Object Detection (Mono 3Det).

\begin{figure*}[t] 
  \centering
  \includegraphics[width=1.0\linewidth]{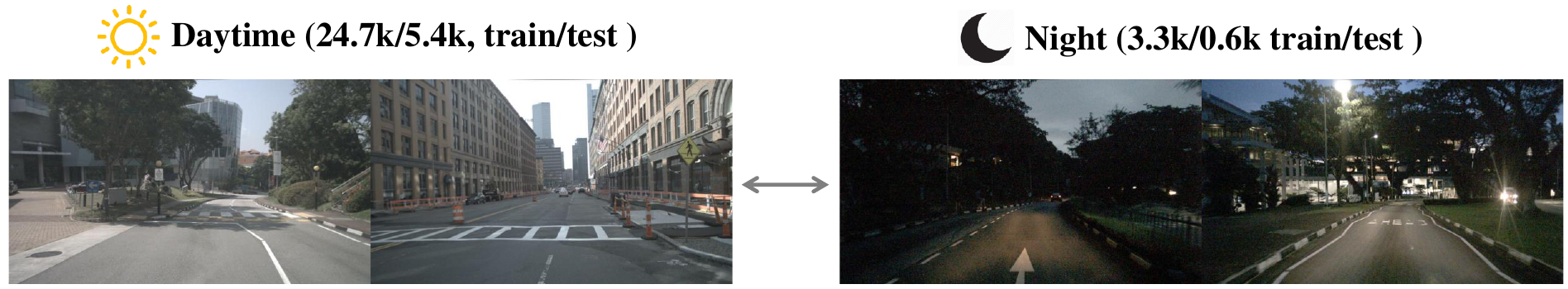} 
   \caption{
   An illustration of the Daytime and Night scenarios of the nuScenes dataset.
   }
   \label{supp:fig_nus}
\end{figure*}

As for two real scenarios of the nuScenes dataset, we first adopt all \emph{front-view} images of nuScenes and convert them into the KITTI format by the official devkit~\cite{caesar2020nuscenes}. Then, we split the adopted images into Daytime or Night scenarios according to their descriptions, following~\cite{liu2023bevfusion}. 
For each scenario, we utilize the images of the training set to achieve a well-trained model and verify the model on the images of the validation set. To be specific, there are 24,745 training, 5,417 test images in the Daytime scenario and 3,385 training, 602 test images in the Night scenario as shown in Fig.~\ref{supp:fig_nus}. Since the nuScenes dataset only contains less than 4k images in the Night scenario with few objects (\eg pedestrians), we only consider the performance of the \emph{Car} category in this dataset.

\section{More Implementation Details}\label{supp:sec:more_impl}
Based on PyTorch~\cite{paszke2019pytorch}, we conduct experiments with NVIDIA A100 (40GB of memory) GPUs and each method is executed on an individual GPU. 
Specifically, we set a learning rate that is half of the original rate employed in the base training for all methods (\ie TENT~\cite{wang2021tent}, EATA~\cite{niu2022efficient} and \ournet), fixing the learning rate throughout the experiment since the optimization of the pre-trained model is restricted to a single epoch.
Furthermore, only the parameters of the \emph{batch normalization layers} are updated. 
It is worth mentioning that we do not employ any form of image augmentation techniques to the test images during model adaptation (\eg rotation and flipping) since it is important to conserve computational resources during the test phase.  

For Monoflex~\cite{zhang2021objects} and MonoGround~\cite{qin2022monoground}, we utilize a randomly generated seed according to their original setting. Both Monoflex and MonoGround employ the same modified DLA-34~\cite{yu2018deep} as the backbone network, with the same input image size of $384 \times 1280$ for the KITTI-C and $928 \times 1600$ for nuScenes, respectively. As for MonoNeRD~\cite{xu2023mononerd}
, it utilizes a lightweight backbone, \ie ResNet34~\cite{he2016deep} with a fixed image resolution of $320 \times 1248$ for the KITTI-C dataset.

\begin{figure*}[t] 
  \centering
  \includegraphics[width=0.9\linewidth]{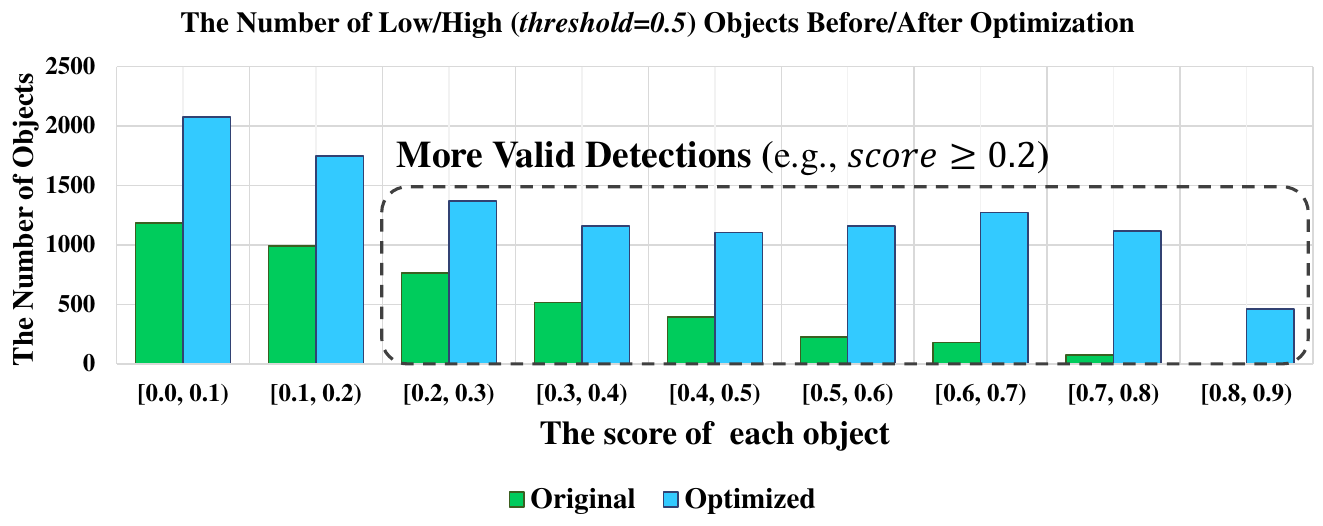} 
  \vspace{-0.1in}
   \caption{
   We further visualize the detailed number of objects within different score margins before and after optimization.
   }
   \label{supp:fig_valid}
\end{figure*}

\begin{table}[t]
\setlength\tabcolsep{10pt}
\renewcommand\arraystretch{0.9}
    \begin{center}
    \caption{\label{tab:nus_res} Comparison with baselines on \textbf{D}aytime $\rightarrow$ \textbf{N}ight and \textbf{N}ight $\rightarrow$ \textbf{D}aytime of {nuScenes}, regarding {Mean $AP_{3D|R_{40}}$}. The \textbf{bold} number indicates the best result.}
        \scalebox{0.6}{
         \begin{tabular}{l|ccc|cccccccc}
         \toprule
         \multirow{2}{*}{Method} & 
         \multicolumn{3}{c|}{MonoFlex} &
         \multicolumn{3}{c}{MonoGround}
         \\
         \cmidrule(lr){2-4} \cmidrule(lr){5-7} 
         & D $\rightarrow$ N & N $\rightarrow$ D & Avg. & D $\rightarrow$ N & N $\rightarrow$ D & Avg. \\
         \midrule
         Pre-trained model & 1.04 & 4.28 & 2.66 & 1.54 & 3.28 & 2.41  \\
         ~~$\bullet~$ BN adaptation~\cite{schneider2020improving} &  7.03 & 4.60 & 5.81 & 5.39 & 4.29 & 4.84 \\
         ~~$\bullet~$ TENT~\cite{wang2021tent} &  10.97 & 1.69 & 6.33 & 6.58 & 3.24 & 4.91   \\
         ~~$\bullet~$ EATA~\cite{niu2022efficient} &  12.31 & 3.53 & 7.92 & 5.43 & 12.16 & 8.79   \\
         ~~$\bullet~$ \ournet~(Ours)  &  \textbf{13.84} & \textbf{4.74} & \textbf{9.29} & \textbf{7.40} & \textbf{13.94} & \textbf{10.67}   \\
         \bottomrule
         \end{tabular}
         }
    \end{center}
    \vspace{-0.15in}
\end{table}

\section{More Experimental Results}\label{supp:sec:more_res}

In this part, we first provide more details of Fig.\textcolor{red}{\textbf{3}}~(b) in the main paper.
As shown in Fig.~\ref{supp:fig_valid}, even though only high-score objects are optimized, the model treats low-score and high-score objects with more confidence, achieving more valid detections. This analysis illustrates the feasibility of our adaptive optimization loss $\mathcal{L}_{AO}$ which aims to conduct reliable model adaptation for OOD test data based on dependable high-score objects.
Then, considering that we visualize the experimental results of nuScenes in the main paper, we provide their detailed results as shown in Table~\ref{tab:nus_res}. It shows \ournet~achieves the best performance within all settings, verifying the effectiveness of our method.
Meanwhile, since we also visualize the experimental results in severity level 2 of KITTI-C in the main paper, we further provide the detailed results as shown in Table~\ref{tab:level_2}. Experimental results at severity level 2 present similar observations to the main paper: 
1) The escalation of severity level results in a larger performance decline within various corruption, enlarging the difficulty of TTA.
2) The performance improvements of existing TTA methods become relatively limited.
3) Even though the task is more challenging, \ournet~still stably obtains the best or comparable performance within all corruption, which demonstrates the superiority of \ournet.

More detailed results based on the image-level inference method MonoNeRD in both severity level 1 and severity level 2 as shown in Table~\ref{tab:nerd_detail1} and Table~\ref{tab:nerd_detail2}. To be specific, the tables show that \ournet~achieves the best or comparable performance under all corruption compared with TENT~\cite{wang2021tent} and EATA~\cite{niu2022efficient}, which demonstrates the effectiveness of the proposed method. Meanwhile, we provide the detailed experimental results of the ablation studies as shown in Table~\ref{tab:ablation_detail} and Table~\ref{tab:ablation_detail2}. Compared with the source-only setting, introducing $\mathcal{L}_{AO}$ or $\mathcal{L}_{Nreg}$ can effectively enhance the model performance. The proposed \ournet~achieves best performance when we combine the adaptive optimization loss $\mathcal{L}_{AO}$ and the negative regularization term $\mathcal{L}_{Nreg}$ together.

\begin{figure*}[t] 
  \centering
  \includegraphics[width=0.9\linewidth]{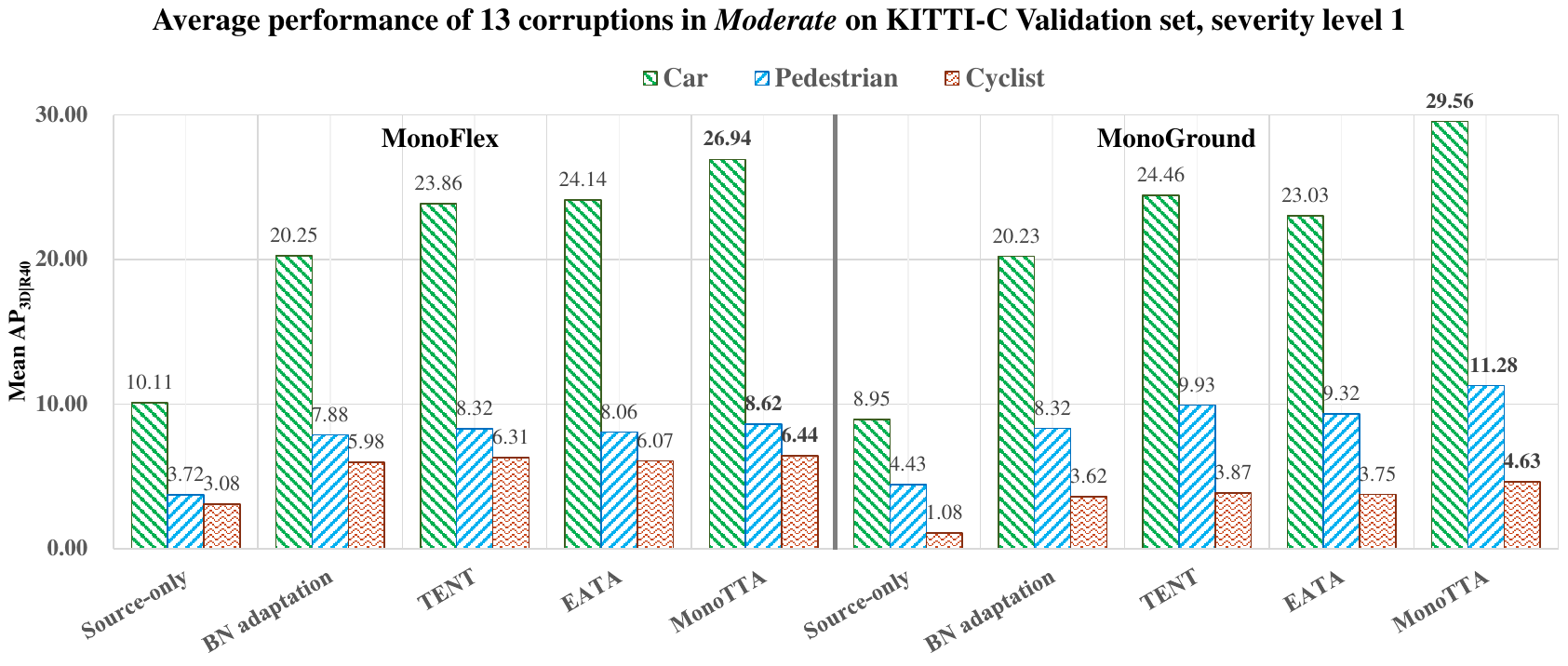} 
  \vspace{-0.1in}
   \caption{
    We further visualize the comparison with baselines on the {KITTI-C validation} set, severity \textbf{level 1} regarding Mean $AP_{3D|R_{40}}$ of the \emph{Moderate} difficulty level. The \textbf{bold} number indicates the best result.
   }
   \label{supp:fig_moderate}
\end{figure*}

In Mono 3Det, the performance for the \emph{Moderate} difficulty level of KITTI-C is one of the most significant indicators of model effectiveness. To this end, we provide more results in severity level 1 and the visualization as shown in Table~\ref{tab:level_1_moderate} and Fig.~\ref{supp:fig_moderate} for \emph{Moderate} (instead of the mean value of \emph{Easy, Moderate, Hard}). The experimental results clearly show that \ournet~maintains the best average performance within all corruptions in terms of Mean $AP_{3D|R_{40}}$ of the \emph{Moderate} difficulty level, which demonstrates the effectiveness and superiority of the proposed \ournet.

\noindent{\textbf{Discussions on High Severity Level Corruption.}}
In this part, we provide more discussions surrounding high-severity data corruptions (\ie3, 4 and 5) based on the experimental results shown in Table~\ref{tab:level_3}, Table~\ref{tab:level_4} and Table~\ref{tab:level_5}. The observations are as follows:
1) With the increase of the severity level, it tends to be more difficult for Fully Test-Time Adaptation (Fully TTA) methods to conduct real-time adaptation due to the further performance degradation of the pre-trained model on test data. For instance, the pre-trained model of MonoGround only achieves 0.60, 0.31 and 0.27 regarding mean $AP_{3D|R_{40}}$ at level 5.
2) Even if our \ournet~still maintains the best performance in these extremely difficult scenarios, all Fully TTA methods only gain limited performance improvement, especially for the Cyclist category. This phenomenon indicates the substantial disparity remaining between the reliability of applications in real applications against ideal conditions.
3) However, as shown in Figure~\ref{supp:fig:level5}, it illustrates that scenarios with high severity levels of corruption are sometimes too difficult even for human beings to recognize every object (\eg defocus and contrast), suggesting that more efforts are necessary to be paid to the research of monocular 3D tasks within extreme corruption. 
To this end, it may be better to utilize additional information like Lidar or multi-view images to enhance the reliability, offering strong motivations for the devising of TTA methods within severe corruption.

\begin{table*}[t]
\setlength\tabcolsep{6pt}
\renewcommand\arraystretch{0.9}
    \begin{center}
    \caption{\label{tab:level_2} 
    Comparison with baselines on the {KITTI-C Validation} set, severity \textbf{level 2} regarding Mean $AP_{3D|R_{40}}$. The \textbf{bold} number indicates the best result.}
    \vspace{-0.05in}
    \scalebox{0.55}{
\begin{tabular}{l|ccc|ccc|cccc|ccc|c}
         \toprule
         \multicolumn{15}{c}{\textbf{\textbf{Car}, IoU=0.5}} \\
         \midrule
         \multirow{2}{*}{Method}  &   
         \multicolumn{3}{c|}{Noise} & 
         \multicolumn{3}{c|}{Blur} & 
         \multicolumn{4}{c|}{Weather} & 
         \multicolumn{3}{c|}{Digital} & 
         \multirow{2}{*}{Avg.} \\
        \cmidrule(lr){2-4} \cmidrule(lr){5-7} \cmidrule(lr){8-11} \cmidrule(lr){12-14} 
        & Gauss. & Shot & Impul. & Defoc. & Glass & Motion & Snow & Frost & Fog & Brit. & Contr. & Pixel & Sat. \\
         \midrule
        Monoflex & 1.24  & 1.28  & 0.50  & 0.61  & 0.64  & 3.10  & 0.17  & 2.81  & 5.52  & 12.38  & 4.23  & 16.27  & 27.78  & 5.89  \\
        ~~$\bullet~$ BN adaptation~\cite{schneider2020improving} & 8.94  & 14.44  & 12.45  & 8.35  & 2.26  & 16.12  & 2.24  & 12.78  & 22.88  & 26.77  & 27.16  & 37.98  & 33.54  & 17.38  \\
         ~~$\bullet~$ TENT~\cite{wang2021tent} &  12.05  & 19.22  & 14.97  & 11.38  & 2.96  & 20.46  & 2.84  & 16.24  & 29.40  & 31.84  & 33.36  & 39.00  & 36.13  & 20.76   \\
         ~~$\bullet~$ EATA~\cite{niu2022efficient} &  12.40  & 19.05  & 16.28  & 13.56  & 4.40  & 21.51  & 3.73  & 17.31  & 29.28  & 31.73  & 33.04  & 32.44  & \textbf{36.91}  & 20.90  \\
         ~~$\bullet~$ \ournet~(Ours) &  \textbf{16.33}  & \textbf{22.32}  & \textbf{19.29}  & \textbf{20.12}  & \textbf{10.15}  & \textbf{26.23}  & \textbf{6.30}  & \textbf{20.79}  & \textbf{32.34}  & \textbf{33.93}  & \textbf{35.39}  & \textbf{39.10}  & 36.84  & \textbf{24.55}  \\
\bottomrule
         \end{tabular}
         }
    \end{center}
    \vspace{-0.15in}
\end{table*}

\begin{table*}[t]
\setlength\tabcolsep{3pt}
\renewcommand\arraystretch{0.95}
    \begin{center}
    \caption{\label{tab:nerd_detail1} 
    MonoNeRD on the {KITTI-C Val} set, severity \textbf{level 1} regarding Mean $AP_{3D|R_{40}}$. The \textbf{bold} number indicates the best result.}
 
    \scalebox{0.65}{
         \begin{tabular}{l|ccc|ccc|cccc|ccc|c}
         \toprule
         \multicolumn{15}{c}{\textbf{\textbf{Car}, IoU=0.5}} \\
         \midrule
         \multirow{2}{*}{Method}  &   
         \multicolumn{3}{c|}{Noise} & 
         \multicolumn{3}{c|}{Blur} & 
         \multicolumn{4}{c|}{Weather} & 
         \multicolumn{3}{c|}{Digital} & 
         \multirow{2}{*}{Avg.} \\
        \cmidrule(lr){2-4} \cmidrule(lr){5-7} \cmidrule(lr){8-11} \cmidrule(lr){12-14} 
        & Gauss. & Shot & Impul. & Defoc. & Glass & Motion & Snow & Frost & Fog & Brit. & Contr. & Pixel & Sat. \\
         \midrule
        MonoNeRD &  14.50 & 27.46 & 17.97 & 8.94 & 18.30 & 20.05 & 2.62 & 11.97 & 9.67 & 29.56 & 9.56 & \textbf{47.11} & 40.20 & 19.84  \\
        ~~$\bullet~$ BN adaptation~\cite{schneider2020improving} & 22.58 & 32.43 & 27.60 & 27.85 & 26.34 & 31.65 & 17.89 & 25.54 & 28.23 & 36.22 & 37.27 & 46.48 & 39.46 & 30.73  \\
         ~~$\bullet~$ TENT~\cite{wang2021tent} &  31.87 & 37.01 & 35.12 & 31.00 & 31.17 & 36.95 & 26.43 & 31.74 & 35.30 & \textbf{40.61} & 40.25 & 46.04 & 40.92 & 35.72   \\
         ~~$\bullet~$ EATA~\cite{niu2022efficient} & 28.31 & 36.80 & 32.29 & 31.40 & 32.26 & 34.72 & 25.56 & 29.06 & 34.88 & 38.65 & 39.05 & 46.30 & 40.57 & 34.60  \\
         ~~$\bullet~$ MonoSSR (Ours) &  \textbf{33.54} & \textbf{40.12} & \textbf{36.81} & \textbf{33.95} & \textbf{35.29} & \textbf{37.14} & \textbf{27.50} & \textbf{33.26} & \textbf{40.29} & 39.91 & \textbf{41.11} & \textbf{45.26} & \textbf{42.00} & \textbf{37.40}   \\
         \bottomrule
         \end{tabular}
         }
    \end{center}
\end{table*}

\begin{table*}[h]
\setlength\tabcolsep{3pt}
\renewcommand\arraystretch{0.95}
    \begin{center}
    \caption{\label{tab:nerd_detail2} 
    MonoNeRD on the {KITTI-C Val} set, severity \textbf{level 2} regarding Mean $AP_{3D|R_{40}}$. The \textbf{bold} number indicates the best result.}
 
    \scalebox{0.65}{
         \begin{tabular}{l|ccc|ccc|cccc|ccc|c}
         \toprule
         \multicolumn{15}{c}{\textbf{\textbf{Car}, IoU=0.5}} \\
         \midrule
         \multirow{2}{*}{Method}  &   
         \multicolumn{3}{c|}{Noise} & 
         \multicolumn{3}{c|}{Blur} & 
         \multicolumn{4}{c|}{Weather} & 
         \multicolumn{3}{c|}{Digital} & 
         \multirow{2}{*}{Avg.} \\
        \cmidrule(lr){2-4} \cmidrule(lr){5-7} \cmidrule(lr){8-11} \cmidrule(lr){12-14} 
        & Gauss. & Shot & Impul. & Defoc. & Glass & Motion & Snow & Frost & Fog & Brit. & Contr. & Pixel & Sat. \\
         \midrule
        MonoNeRD & 6.71 & 14.66 & 7.16 & 2.51 & 6.83 & 9.15 & 0.34 & 3.84 & 6.96 & 20.26 & 5.14 & \textbf{45.82} & 39.83 & 13.02   \\
        ~~$\bullet~$ BN adaptation~\cite{schneider2020improving} & 17.64 & 26.57 & 20.63 & 22.13 & 17.48 & 27.04 & 13.40 & 16.92 & 26.29 & 33.80 & 36.88 & 45.45 & 39.91 & 26.47  \\
         ~~$\bullet~$ TENT~\cite{wang2021tent} &  24.40 & 35.95 & 26.94 & 29.74 & 23.62 & 32.40 & 17.42 & 24.14 & 34.22 & 38.94 & 40.01 & 45.19 & 41.02 & 31.85   \\
         ~~$\bullet~$ EATA~\cite{niu2022efficient} & 24.05 & 33.18 & 26.92 & 26.49 & 23.94 & 29.81 & 17.94 & 21.19 & 33.23 & 36.26 & 38.98 & 45.81 & 40.74 & 30.66  \\
         ~~$\bullet~$ MonoSSR (Ours) &  \textbf{28.50} & \textbf{36.12} & \textbf{28.96} & \textbf{30.26} & \textbf{29.61} & \textbf{34.08} & \textbf{22.95} & \textbf{24.94} & \textbf{38.56} & \textbf{40.17} & \textbf{41.02} & 45.40 & \textbf{41.35} & \textbf{33.99}   \\
         \bottomrule
         \end{tabular}
         }
    \end{center}
\end{table*}

\begin{table*}[t]
\setlength\tabcolsep{3pt}
    \begin{center}
    \caption{\label{tab:ablation_detail}
    Based on MonoGround~\cite{qin2022monoground}, we conduct ablation studies of $\mathcal{L}_{AO}$ and $\mathcal{L}_{Nreg}$ on the level-1 {KITTI-C Val} set, regarding {$AP_{3D|R_{40}}$}.
    }
    \scalebox{0.65}{
         \begin{tabular}{ccc|ccc|ccc|cccc|ccc|c}
         \toprule
         \multicolumn{15}{c}{\textbf{\textbf{Car}, IoU=0.5}} \\
         \midrule
         \multirow{2}{*}{Backbone} & \multirow{2}{*}{$\mathcal{L}_{AO}$} & \multirow{2}{*}{$\mathcal{L}_{Nreg}$} & 
         \multicolumn{3}{c|}{Noise} & 
         \multicolumn{3}{c|}{Blur} & 
         \multicolumn{4}{c|}{Weather} & 
         \multicolumn{3}{c|}{Digital} & 
         \multirow{2}{*}{Avg.} \\
         \cmidrule(lr){4-6} \cmidrule(lr){7-9} \cmidrule(lr){10-13} \cmidrule(lr){14-16}
         & & & Gauss. & Shot & Impul. & Defoc. & Glass & Motion & Snow & Frost & Fog & Brit. & Contr. & Pixel & Sat. \\
         \midrule
         \cmark &  &  &  2.40 & 4.10 & 3.31 & 3.71 & 2.67 & 8.13 & 0.22 & 5.54 & 4.59 & 25.37 & 4.00 & 33.57 & 28.08 & 9.67  \\
         \cmark & \cmark &  & 18.43  & 27.99  & 23.89  & 22.71  & 22.67  & 29.37  & 14.29  & 26.49  & 31.25  & 33.68  & 34.30  & 43.39  & 34.77  & 27.94  \\
         \cmark &  & \cmark &  8.18  & 23.73  & 20.24  & 20.85  & 8.77  & 24.15  & 8.19  & 20.94  & 25.16  & 31.91  & 31.80  & 43.92  & 34.51  & 23.26  \\
         \cmark & \cmark & \cmark &  26.13  & 33.11  & 22.54  & 30.38  & 25.48  & 32.44  & 18.72  & 32.60  & 37.75  & 37.87  & 39.57  & 43.67  & 37.98  & 32.17  \\
                  \toprule
         \multicolumn{15}{c}{\textbf{\textbf{Pedestrian}, IoU=0.25}} \\
\bottomrule
         \end{tabular}
         }
    \end{center}
\end{table*}

\begin{table*}[t]
\setlength\tabcolsep{3pt}
    \begin{center}
    \caption{\label{tab:ablation_detail2}
    Based on MonoGround~\cite{qin2022monoground}, we conduct ablation studies of $\mathcal{L}_{AO}$ and $\mathcal{L}_{Nreg}$ on the level-2 {KITTI-C Val} set, regarding {$AP_{3D|R_{40}}$}.
    }
    \scalebox{0.65}{
         \begin{tabular}{ccc|ccc|ccc|cccc|ccc|c}
         \toprule
         \multicolumn{15}{c}{\textbf{\textbf{Car}, IoU=0.5}} \\
         \midrule
         \multirow{2}{*}{Backbone} & \multirow{2}{*}{$\mathcal{L}_{AO}$} & \multirow{2}{*}{$\mathcal{L}_{Nreg}$} & 
         \multicolumn{3}{c|}{Noise} & 
         \multicolumn{3}{c|}{Blur} & 
         \multicolumn{4}{c|}{Weather} & 
         \multicolumn{3}{c|}{Digital} & 
         \multirow{2}{*}{Avg.} \\
         \cmidrule(lr){4-6} \cmidrule(lr){7-9} \cmidrule(lr){10-13} \cmidrule(lr){14-16}
         & & & Gauss. & Shot & Impul. & Defoc. & Glass & Motion & Snow & Frost & Fog & Brit. & Contr. & Pixel & Sat. \\
         \midrule
         \cmark &  &  &  0.53  & 0.73  & 0.44  & 1.89  & 1.05  & 1.73  & 0.43  & 1.63  & 2.80  & 12.96  & 1.66  & 26.40  & 22.09  & 5.72   \\
         \cmark & \cmark &  &  11.60  & 19.30  & 17.45  & 15.27  & 8.33  & 8.68  & 9.71  & 16.86  & 28.64  & 30.10  & 32.28  & 41.92  & 36.36  & 21.27  \\
         \cmark &  & \cmark &  8.49  & 16.51  & 13.27  & 9.65  & 2.81  & 17.95  & 4.14  & 22.52  & 22.45  & 26.44  & 26.34  & 41.38  & 34.20  & 18.93   \\
         \cmark & \cmark & \cmark &  15.50  & 25.39  & 21.97  & 20.93  & 13.56  & 28.58  & 5.42  & 22.45  & 32.54  & 32.85  & 35.19  & 41.93  & 37.31  & 25.66   \\
        \toprule
         \multicolumn{15}{c}{\textbf{\textbf{Pedestrian}, IoU=0.25}} \\
\bottomrule
         \end{tabular}
         }
    \end{center}
\end{table*}

\begin{table*}[t]
\setlength\tabcolsep{8pt}
\renewcommand\arraystretch{0.9}
    \begin{center}
    \caption{\label{tab:level_1_moderate} 
    Comparisons on the {KITTI-C Validation} set, severity {level 1} regarding $AP_{3D|R_{40}}$ of the \emph{Moderate} difficulty level. The \textbf{bold} number indicates the best result.}
    \vspace{-0.05in}
    \scalebox{0.5}{
         \begin{tabular}{l|ccc|ccc|cccc|ccc|c}
         \toprule
        \multicolumn{15}{c}{\textbf{\textbf{Car}, IoU=0.25}} \\
         \midrule
         \multirow{2}{*}{Method}  &   
         \multicolumn{3}{c|}{Noise} & 
         \multicolumn{3}{c|}{Blur} & 
         \multicolumn{4}{c|}{Weather} & 
         \multicolumn{3}{c|}{Digital} & 
         \multirow{2}{*}{Avg.} \\
        \cmidrule(lr){2-4} \cmidrule(lr){5-7} \cmidrule(lr){8-11} \cmidrule(lr){12-14} 
        & Gauss. & Shot & Impul. & Defoc. & Glass & Motion & Snow & Frost & Fog & Brit. & Contr. & Pixel & Sat. \\
         \midrule
        Monoflex & 3.48 & 6.40 & 4.77 & 2.02 & 3.38 & 10.09 & 0.26 & 7.41 & 7.46 & 23.87 & 6.95 & 26.09 & 29.26 & 10.11\\
         ~~$\bullet~$ BN adaptation~\cite{schneider2020improving} & 12.55 & 19.53 & 17.14 & 14.49 & 7.72 & 22.37 & 5.22 & 19.83 & 22.43 & 29.25 & 27.65 & 37.56 & 27.57 & 20.25  \\
         ~~$\bullet~$ TENT~\cite{wang2021tent} &   16.20 & 24.15 & 20.98 & 19.82 & 10.56 & 26.32 & 6.40 & 24.25 & 27.52 & 32.29 & 31.52 & 37.98 & 32.23 & 23.86   \\
         ~~$\bullet~$ EATA~\cite{niu2022efficient} & 15.49 & 23.76 & 22.69 & 20.32 & 12.07 & 24.89 & 7.41 & 24.17 & 28.43 & 32.11 & 31.72 & 38.02 & 32.75 & 24.14  \\
         ~~$\bullet~$ \ournet~(Ours) & \textbf{19.32} & \textbf{25.88} & \textbf{24.43} & \textbf{23.83} & \textbf{17.33} & \textbf{28.63} & \textbf{11.39} & \textbf{27.81} & \textbf{30.72} & \textbf{33.73} & \textbf{33.65} & \textbf{38.94} & \textbf{34.56} & \textbf{26.94}  \\
\bottomrule
         \end{tabular}
         }
    \end{center}
    \vspace{-0.1in}
\end{table*}

\begin{table*}[t]
\setlength\tabcolsep{8pt}
\renewcommand\arraystretch{0.9}
    \begin{center}
    \caption{\label{tab:level_3} 
    Comparison on the {KITTI-C Val} set, severity \textbf{level 3} regarding Mean $AP_{3D|R_{40}}$. The \textbf{bold} number indicates the best result.}
 
    \scalebox{0.5}{
         \begin{tabular}{l|ccc|ccc|cccc|ccc|c}
         \toprule
         \multicolumn{15}{c}{\textbf{\textbf{Car}, IoU=0.5}} \\
         \midrule
         \multirow{2}{*}{Method}  &   
         \multicolumn{3}{c|}{Noise} & 
         \multicolumn{3}{c|}{Blur} & 
         \multicolumn{4}{c|}{Weather} & 
         \multicolumn{3}{c|}{Digital} & 
         \multirow{2}{*}{Avg.} \\
        \cmidrule(lr){2-4} \cmidrule(lr){5-7} \cmidrule(lr){8-11} \cmidrule(lr){12-14} 
        & Gauss. & Shot & Impul. & Defoc. & Glass & Motion & Snow & Frost & Fog & Brit. & Contr. & Pixel & Sat. \\
         \midrule
        Monoflex &  0.84 & 0.54 & 0.62 & 0.64 & 0.00 & 0.48 & 0.13 & 1.70 & 4.07 & 6.41 & 1.98 & 11.79 & 23.96 & 4.09  \\
        ~~$\bullet~$ BN adaptation~\cite{schneider2020improving} & 5.14 & 9.47 & 8.68 & 2.85 & \textbf{0.14} & 8.67 & 2.87 & 10.02 & 20.63 & 21.86 & 20.44 & 35.05 & 27.40 & 13.32  \\
         ~~$\bullet~$ TENT~\cite{wang2021tent} &  6.86 & 13.07 & 11.80 & 3.43 & 0.12 & 11.01 & 3.04 & 11.71 & 27.34 & 27.40 & 26.73 & 36.53 & 32.26 & 16.25   \\
         ~~$\bullet~$ EATA~\cite{niu2022efficient} & 7.13 & 13.94 & 12.23 & 3.72 & 0.09 & 12.27 & 3.90 & 13.27 & 26.96 & 27.38 & 27.29 & \textbf{36.83} & 31.78 & 16.68  \\
         ~~$\bullet~$ MonoSSR (Ours) &  \textbf{9.74} & \textbf{16.89} & \textbf{15.28} & \textbf{8.32} & 0.03 & \textbf{19.25} & \textbf{4.76} & \textbf{16.83} & \textbf{31.09} & \textbf{29.88} & \textbf{30.63} & 36.45 & \textbf{33.69} & \textbf{19.45}   \\
\bottomrule
         \end{tabular}
         }
    \end{center}
    \vspace{-0.1in}
\end{table*}

\begin{table*}[t]
\setlength\tabcolsep{8pt}
\renewcommand\arraystretch{0.9}
    \begin{center}
    \caption{\label{tab:level_4} 
    Comparison on the {KITTI-C Val} set, severity \textbf{level 4} regarding Mean $AP_{3D|R_{40}}$. The \textbf{bold} number indicates the best result.}
 
    \scalebox{0.5}{
         \begin{tabular}{l|ccc|ccc|cccc|ccc|c}
         \toprule
         \multicolumn{15}{c}{\textbf{\textbf{Car}, IoU=0.5}} \\
         \midrule
         \multirow{2}{*}{Method}  &   
         \multicolumn{3}{c|}{Noise} & 
         \multicolumn{3}{c|}{Blur} & 
         \multicolumn{4}{c|}{Weather} & 
         \multicolumn{3}{c|}{Digital} & 
         \multirow{2}{*}{Avg.} \\
        \cmidrule(lr){2-4} \cmidrule(lr){5-7} \cmidrule(lr){8-11} \cmidrule(lr){12-14} 
        & Gauss. & Shot & Impul. & Defoc. & Glass & Motion & Snow & Frost & Fog & Brit. & Contr. & Pixel & Sat. \\
         \midrule
        Monoflex &  0.00  & 0.00  & 0.00  & 0.00  & 0.00  & 0.04  & 0.28  & 1.59  & 3.61  & 2.82  & 0.00  & 4.39  & 1.71  & 1.11  \\
        ~~$\bullet~$ BN adaptation~\cite{schneider2020improving} & 2.99  & 4.64  & 3.82  & 0.52  & 0.04  & 4.18  & 2.21  & 9.34  & 21.52  & 17.75  & 13.98  & 27.15  & 16.32  & 9.57  \\
         ~~$\bullet~$ TENT~\cite{wang2021tent} &  3.78  & 5.85  & 5.87  & 0.56  & 0.04  & 4.50  & 2.27  & 10.53  & 27.66  & 24.02  & 19.78  & 30.44  & 21.78  & 12.08   \\
         ~~$\bullet~$ EATA~\cite{niu2022efficient} & 4.26  & \textbf{6.84}  & 5.80  & 0.61  & \textbf{0.05}  & 5.68  & 2.93  & 12.51  & 27.34  & 24.45  & 19.59  & 30.43  & 22.88  & 12.57  \\
         ~~$\bullet~$ MonoSSR (Ours) &  \textbf{4.77}  & 5.77  & \textbf{7.79} & \textbf{1.40}  & 0.01  & \textbf{11.71}  & \textbf{4.23}  & \textbf{16.66}  & \textbf{31.34}  & \textbf{27.67}  & \textbf{23.65}  & \textbf{30.95}  & \textbf{25.97}  & \textbf{14.76}   \\
\bottomrule
         \end{tabular}
         }
    \end{center}
\end{table*}

\begin{table*}[t]
\setlength\tabcolsep{8pt}
\renewcommand\arraystretch{0.9}
    \begin{center}
    \caption{\label{tab:level_5} 
    Comparison on the {KITTI-C Val} set, severity \textbf{level 5} regarding Mean $AP_{3D|R_{40}}$. The \textbf{bold} number indicates the best result.}
 
    \scalebox{0.5}{
         \begin{tabular}{l|ccc|ccc|cccc|ccc|c}
         \toprule
         \multicolumn{15}{c}{\textbf{\textbf{Car}, IoU=0.5}} \\
         \midrule
         \multirow{2}{*}{Method}  &   
         \multicolumn{3}{c|}{Noise} & 
         \multicolumn{3}{c|}{Blur} & 
         \multicolumn{4}{c|}{Weather} & 
         \multicolumn{3}{c|}{Digital} & 
         \multirow{2}{*}{Avg.} \\
        \cmidrule(lr){2-4} \cmidrule(lr){5-7} \cmidrule(lr){8-11} \cmidrule(lr){12-14} 
        & Gauss. & Shot & Impul. & Defoc. & Glass & Motion & Snow & Frost & Fog & Brit. & Contr. & Pixel & Sat. \\
         \midrule
        Monoflex & 0.00 & 0.00 & 0.00 & 0.00 & 0.00 & 0.00 & 0.42 & 0.78 & 2.84 & 1.13 & 0.00 & 2.12 & 0.62 & 0.61   \\
        ~~$\bullet~$ BN adaptation~\cite{schneider2020improving} & 1.35 & 2.50 & 1.85 & \textbf{0.25} & \textbf{0.04} & 2.56 & 2.19 & 7.44 & 19.94 & 13.75 & 9.34 & 22.84 & 11.43 & 7.34  \\
         ~~$\bullet~$ TENT~\cite{wang2021tent} &  1.71 & 2.93 & 2.31 & 0.17 & \textbf{0.04} & 2.56 & 2.89 & 8.68 & 25.61 & 19.02 & 12.88 & 25.04 & 17.65 & 9.35   \\
         ~~$\bullet~$ EATA~\cite{niu2022efficient} & 1.81 & 3.46 & 2.51 & 0.25 & \textbf{0.04} & 3.14 & 2.91 & 9.26 & 25.72 & 20.85 & 14.85 & 24.92 & 17.79 & 9.81  \\
         ~~$\bullet~$ MonoSSR (Ours) &  \textbf{2.41} & \textbf{4.35} & \textbf{2.04} & 0.03 & 0.01 & \textbf{8.56} & \textbf{5.37} & \textbf{12.18} & \textbf{30.01} & \textbf{23.70} & \textbf{19.91} & \textbf{25.74} & \textbf{22.83} & \textbf{12.09}   \\
\bottomrule
         \end{tabular}
         }
    \end{center}
    \vspace{-0.1in}
\end{table*}

\end{document}